\documentclass[journal]{IEEEtran}
\usepackage{amsmath,amsfonts}
\usepackage{algorithmic}
\usepackage{algorithm}
\usepackage{array}
\usepackage{textcomp}
\usepackage{stfloats}
\usepackage{url}
\usepackage{verbatim}
\usepackage{graphicx}
\usepackage{cite}
\usepackage{customization}

\begin{document}

\title{A-ACT: Action Anticipation through\\ Cycle Transformations}

\author{Akash Gupta, Jingen Liu~\IEEEmembership{Senior Member,~IEEE}, Liefeng Bo,~\IEEEmembership{Member,~IEEE}, \\ Amit K. Roy-Chowdhury,~\IEEEmembership{Fellow,~IEEE} and Tao Mei,~\IEEEmembership{Fellow,~IEEE}% <-this % stops a space
    \thanks{$\bullet$ Akash Gupta, and Amit~K.~Roy-Chowdhury are with the Department of Electrical and Computer Engineering, University of California, Riverside, CA, USA. This work was done when Akash Gupta was an intern at JD.com AI Research. Jingen Liu, Liefeng Bo, and Tao Mei are with JD.com AI Research, Mountain View, CA, USA.  E-mails: (agupt013@ucr.edu, jingenliu@gmail.com,  liefeng.bo@jd.com, tmei@live.com, amitrc@ece.ucr.edu)}}% <-this % stops a space

\markboth{IEEE Transactions on Image Processing,~Vol.xx, No.xx, xxx~2021}%
{Gupta \MakeLowercase{\textit{et al.}}: A-ACT: Action Anticipation through Cycle Transformations}


\maketitle

\begin{abstract}
%%%%%%% Background %%%%%%%
While action anticipation has garnered a lot of research interest recently, most of the works focus on anticipating future action directly through observed visual cues only. 
In this work, we take a step back to analyze how the human capability to anticipate the future can be transferred to machine learning algorithms. 
To incorporate this ability in intelligent systems a question worth pondering upon is how exactly do we anticipate? Is it by anticipating future actions from past experiences? Or is it by simulating possible scenarios based on cues from the present? A recent study on human psychology~\cite{breska2018double} explains that, in anticipating an occurrence, the human brain counts on both systems. 
%%
% Anticipation is 
In this work, we study the impact of each system for the task of action anticipation and  introduce a paradigm to integrate them in a learning framework. We believe that intelligent systems designed by leveraging the psychological anticipation models will do a more nuanced job at the task of human action prediction.
Furthermore, we introduce cyclic transformation in temporal dimension in feature and semantic label space to instill the human ability of reasoning of past actions based on the predicted future. Experiments on Epic-Kitchen, Breakfast, and 50Salads dataset demonstrate that the action anticipation model learned using a combination of the two systems along with the cycle transformation performs favorably against various state-of-the-art approaches.

\end{abstract}


% ------------------ Introduction
\section{Introduction}

Action anticipation is essential for various real-world applications such as autonomous navigation~\cite{bahavan2020anomaly, liu2020spatiotemporal} and assistive robots for human-machine interaction~\cite{kriegel2019socially, argall2015turning, clabaugh2018robots, ke2019learning}. Thus, it is paramount to incorporate anticipation ability in intelligent systems. Recently, some progress has been made to model the anticipation capability and embed it in intelligent and robotic systems using deep learning networks~\cite{liu2020forecasting, sanchez2020hardware, massardi2020parc, saxena2013anticipating, ke2018learning, roy2021action, huang2016deep}. One naive approach to address the task of action anticipation is to learn a direct mapping between observed visual cues and the future action using supervised methods~\cite{simonyan2014two, wang2016temporal}. 
Other approaches translate the observed cues to the future visual cues using pre-computed features~\cite{vondrick2016anticipating, furnari2019would} and then perform action recognition, thereby the task of anticipation. 
However, what comes more naturally to humans is challenging for intelligent systems due to the complexity of the task of anticipation and the stochastic nature of the future.

Anticipation is one of the neuro-cognitive mechanisms of our brain. We constantly try to anticipate what will happen next depending on the knowledge our brain has of the environment. A study on human psychology~\cite{breska2018double} explains that humans count on two systems while preforming the task of anticipation. One system allows us to utilize our semantic experience to anticipate future, while the other is based on identifying the visual patterns.
A skilled cricket batsman can easily visualize the trajectory of the ball by observing the movement \emph{pattern} of the bowler and play a shot. On the other hand, if the same cricket player is asked to play baseball, with little knowledge of the sport, he can apply the \emph{experience} gained through cricket to make a hit. 
While in the former example the player is skilled enough to visualize the future trajectories and make a decision based on other cues to anticipate action, in the latter the player uses the experience gained from another sport to learn and anticipate. This ability to leverage experiences and identify patterns in the present make it feasible for humans to anticipate the future (at least, the near-future).
%
% The workflow resembling the two temporal anticipation mechanisms that assist humans is presented in Figure~\ref{fig:teaser}.
% % Given the observed visual cues $\mathbf{X}_o$ our objective is to anticipate future action $\widehat{\textbf{a}}_f$. To this end, we integrate two anticipation systems, described in the study~\cite{breska2018double}, as modules in our action anticipation model.
% % Before applying these modules we first employ a self-attention based feature translation module $\mathcal{G}_t$ extract intermediate feature representation $\mathbf{X}_t$ to learn the correlated features.
% %
% While one mechanism visualizes the future possibilities $\widehat{\mathbf{X}}_f$, using the observed features $\mathbf{X}_o$, for anticipation as in case of skilled cricketer playing cricket (bottom-branch), the other is based on experiences to understand the current action $\widehat{\textbf{a}}_o$ before anticipating the future, similar to cricket player utilizing their experience when learning to play baseball (top-branch).

%A study on human psychology~\cite{breska2018double} explains that humans count on two system while preforming the task of anticipation. One system, allows us to utilize our past experience to anticipate future, while other is based on identifying the patterns. 
\begin{figure}[t]
    \centering
    \includegraphics[width=0.98\columnwidth]{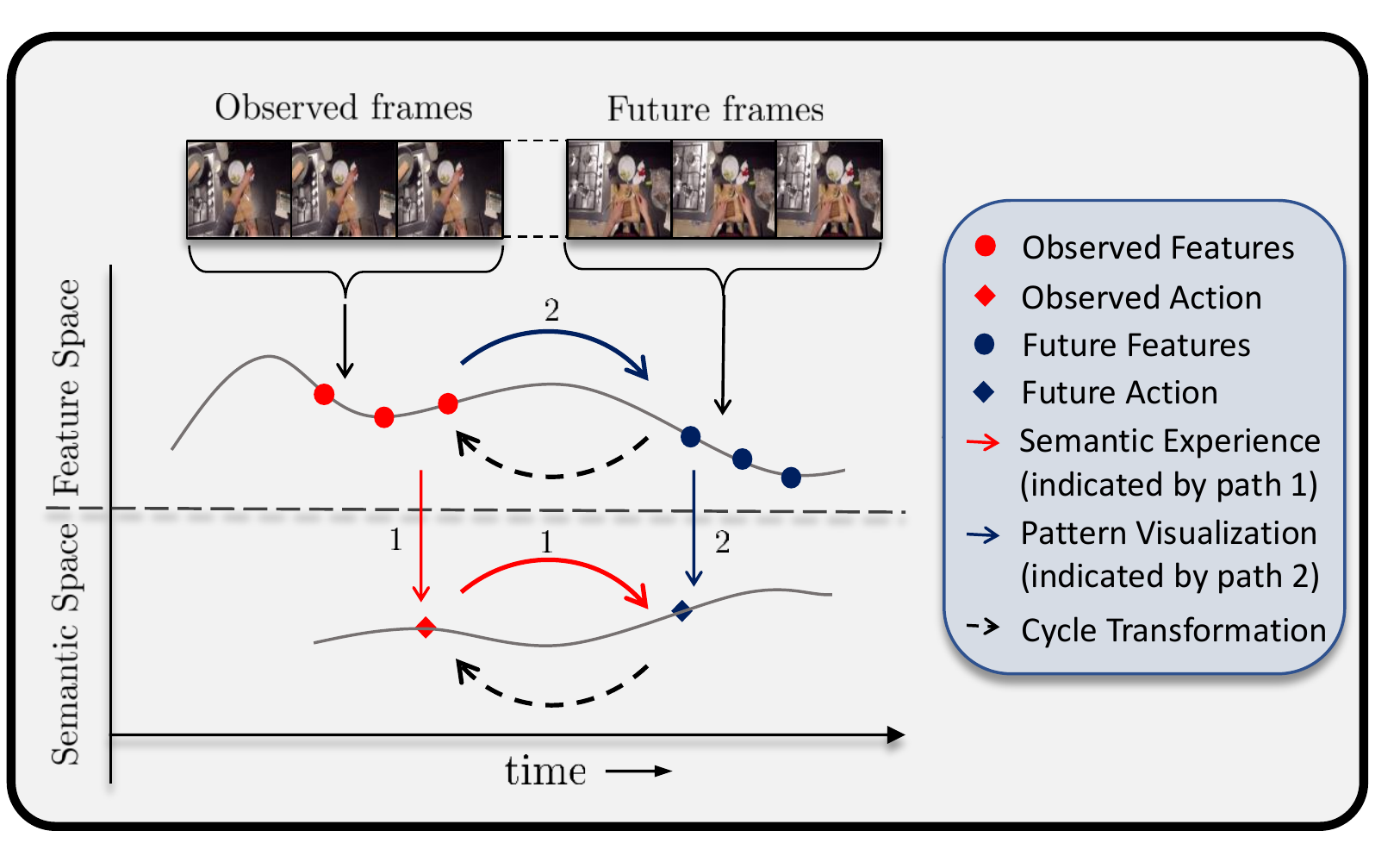}
    % \vspace{-1mm}
    \caption{\textbf{Conceptual Overview.} Our proposed approach \textbf{A-ACT} unifies the semantic past experience mechanism with the pattern visualization. We employ cycle-transformations in the semantic label space (cycle with label 1) as well as the feature space (cycle with label 2) to learn discriminative features. }
    % \label{fig:conceptual_overview}
    \label{fig:teaser}
    
\end{figure}
Motivated by aforementioned cognitive study, we propose a novel approach for \textbf{A}ction \textbf{A}nticipation through \textbf{C}ycle \textbf{T}ransformations (\textbf{A-ACT}). Unlike other approaches ~\cite{simonyan2014two, wang2016temporal, vondrick2016anticipating, furnari2019would, wu2020learning, quan2021holistic} where only one of the anticipation mechanisms is studied, we study the impact of both the mechanisms for action anticipation using deep learning models. 
As an analogy, in challenging scenarios where fine details are needed to anticipate the future, like examining a video of crime scene, we may have to confirm details by rewinding and forwarding the tape to find reasons behind past and current actions. We try to translate this ability to reason in our model through cycle transformation in temporal dimension.
We propose that integrating the psychological philosophy into designing intelligent systems will do a more nuanced job at human action anticipation, as subsequently demonstrated by the results of our experiments.% Additionally, cycle transformation in temporal dimension is utilized to reinforce the ability to reason regarding the past action based on the predicted future. 

%We also employ cycle transformation in the temporal dimension to integrate these modules and enforce additional constraint for feature representations.}

% However, even with the abundance of cues to predict the future, human action anticipation still remains notoriously challenging due to stochastic nature of the future. 
% The two temporal anticipation mechanisms that assist humans are presented in Figure~\ref{fig:teaser}. Given the observed visual cues our objective is to anticipate future action. To this end, we integrate two anticipation systems, described in the study~\cite{breska2018double}, as modules in our action anticipation model. While one module based on past experiences requires to understand the current action before anticipating the future (Figure~\ref{fig:overview} top-block), the other module visualizes the future possibilities in order to anticipate the future (Figure~\ref{fig:overview} bottom-block). We also employ cycle transformation in the temporal dimension to integrate these modules and enforce additional constraint for feature representations.

The conceptual overview of the proposed approach that unifies the semantic experience and patter visualization is shown in Fig.~\ref{fig:teaser}. Given a set of observed frames, we project those frames onto a feature space. For the semantic experience (\textbf{SE}) mechanism, we first recognize the current action and then anticipate the future action from the current action (see Fig.~\ref{fig:teaser}; path 1 in green). Since, the \textbf{SE} model utilizes the infered current action labels for future anticipation, it completely relies on the semantic labels.  On the other hand the pattern visualization (\textbf{PV}) mechanism first generates the probable future features and then performs action anticipation (see Fig.~\ref{fig:teaser}; path 2 in blue). The pattern visualization model \textbf{PV} is trying to find a pattern for future possibilities based on the observed features and utilizes the features pattern for the task of anticipation. 

We enforce the cycle-consistency constrains on the feature space as well as the semantic label space as shown in Fig.~\ref{fig:teaser}. 
The feature cycle-transformation is applied for the patter visualization model. The future features generated using the pattern visualization model is used to reconstruct the observed feature thereby enforcing the cycle-consistency loss in the feature space. On the other hand, the semantic cycle-consistency is applied between the action anticipated using the generated future features in case of the pattern visualization model and action anticipated using the reconstructed observed features in the semantic experience model. Experiments on various dataset show that incorporating the human temporal anticipation mechanism using the cycle-consistency in semantic label as well feature space can help learn the task of action anticipation better.

\begin{figure*}[t]
    \centering
    \includegraphics[width=0.65\textwidth]{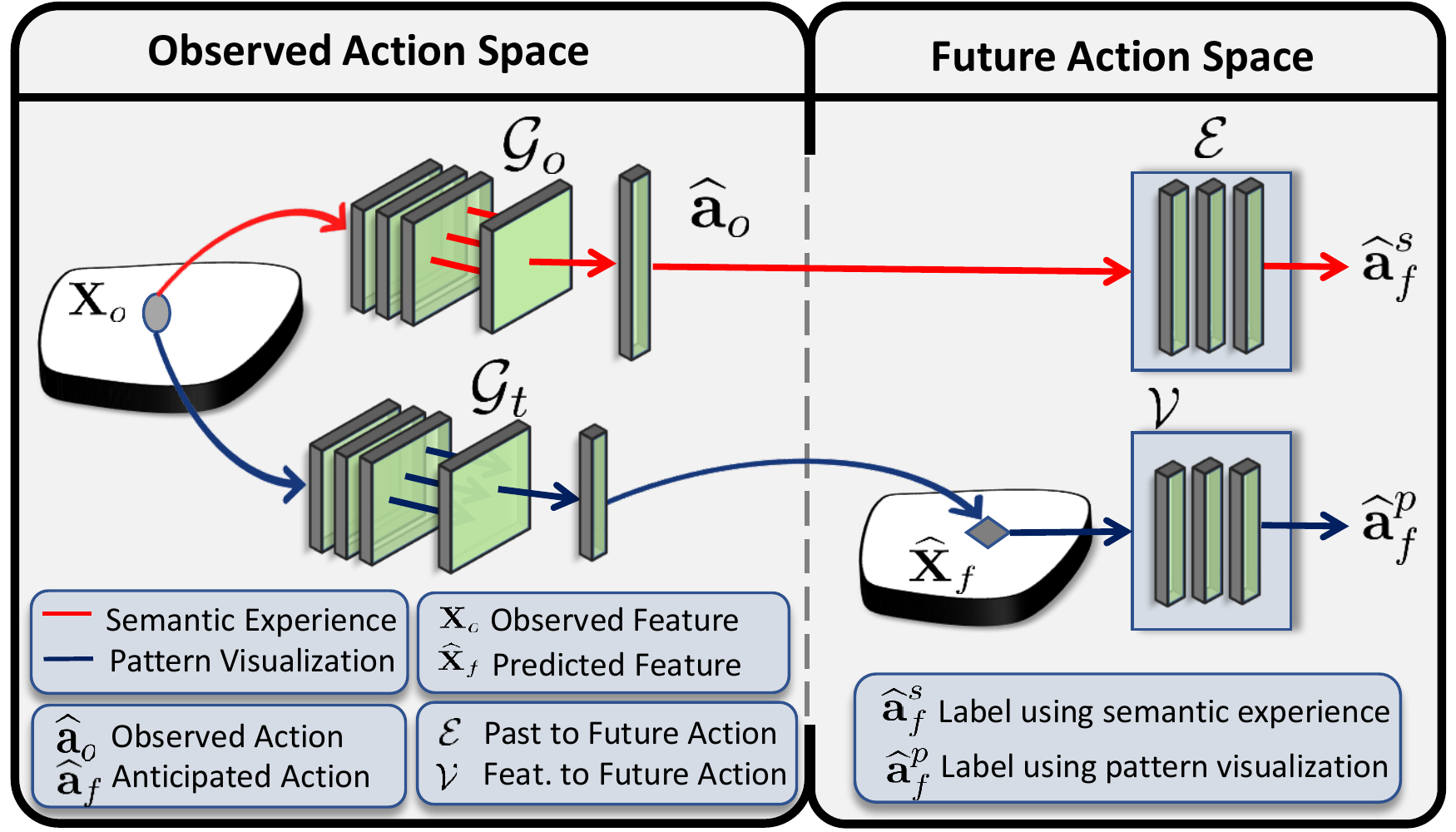}
    \caption{Illustrative methods resembling human temporal prediction mechanisms - semantic experience and visualization of future patterns~\cite{breska2018double}. Given the features of the observed cues, the semantic experience mechanism first identifies the observed action using self-attention module $\mathcal{G}_o$ and then anticipates the future action (top-branch in red; see section~\ref{sec:SE}). Pattern visualization mechanism uses the observed features to generate plausible future features with module $\mathcal{G}_t$ and performs anticipation using these features (bottom-branch in blue; see section~\ref{sec:PV})).} 
    % \amit{Make this 2-column - will be easier to read.}
    % \vspace{-3mm}
    \label{fig:se-pv}
    
\end{figure*}

The working of \textbf{SE} and \textbf{PV}, that resembles two  human  temporal  prediction mechanisms, is presented in Figure~\ref{fig:se-pv}.  Given the features of the observed cues $\mathbf{X}_o$, the semantic experience \textbf{SE} model first identifies the observed action ($\textbf{a}_o$) using self-attention module $\mathcal{G}_o$ and then utilize the anticipation layer $\mathcal{E}$ to anticipate the future action $\textbf{a}_f$ (top-branch in red; see section~\ref{sec:SE}). Pattern visualization \textbf{PV} model uses the observed features $\mathbf{X}_o$ to generate plausible future features $\widehat{\mathbf{X}}_f$ with feature translation module $\mathcal{G}_t$. Then action is anticipated using the model $\mathcal{V}$ on the generated features $\widehat{\mathbf{X}}_f$ for future action $\widehat{\textbf{a}}_f^{~p}$ (bottom-branch in blue; see section~\ref{sec:PV})). 

An overview of our approach \textbf{A-ACT} is illustrated in Figure~\ref{fig:overview}. Our cycle consistency model is composed of two feature translation modules ($\mathcal{G}_{r}$, $\mathcal{G}_{a}$), two recognition modules ($\mathcal{V}_r$, $\mathcal{V}_a$) for past action and future action and an experience model $\mathcal{E}$ for action anticipation using semantic past label. The forward cycle consists of $\mathcal{G}_{a}$ that takes observed features $\mathbf{X}_o$ and translates them into possible future features $\widehat{\mathbf{X}}_f$ and future action recognition module $\mathcal{V}_a$ to anticipate future action $\widehat{\textbf{a}}_f^{~p}$. The reverse cycle utilizes $\mathcal{G}_{r}$ to reconstruct the observed features $\widehat{\mathbf{X}}_o$ from the generated future features $\widehat{\mathbf{X}}_f$ followed by a past action recognition module $\mathcal{V}_r$ to obtain observed action semantic label $\widehat{\textbf{a}}_o$ . The experience model uses the semantic label to anticipate future action $\widehat{\textbf{a}}_f^{~s}$. The cycle transformation is applied between in the semantic label space by minimizing the distribution between action anticipated using semantic experience $\widehat{\textbf{a}}_f^{~s}$ and pattern visualization $\widehat{\textbf{a}}_f^{~p}$.  The cycle transformation in feature space is enforced by minimizing the $\ell_2$ distance between the reconstructed features $\widehat{\mathbf{X}}_o$ and the observed features $\mathbf{X}_o$.\bigskip

The key contributions of our work are as follows.\smallskip

\begin{itemize}
    \setlength \itemsep{0.4em}
    \item We propose an efficient framework that incorporates the psychological study on human anticipation ability 
    to learn discriminative representations 
    for the task of anticipation.
    \item To achieve this, we propose temporal cycle transformations between feature and label space, thus capturing both the semantic experience aspect and pattern visualization aspect of action anticipation.
    \item Experiments on various benchmark datasets demonstrate the proposed approach performs favourably against various state-of-the-art approaches. Furthermore, in ablation study we show that our model preforms well even in limited data setting.
\end{itemize}

% ------------------ Related Work
\section{Related Work}

Our work relates to three major research directions: early action recognition, anticipation and cycle consistency. 
This section focuses on some representative methods closely related to our work.
\medskip

\begin{table}[b]

\caption{\textbf{Categorization of some of the representative methods in action anticipation}. 
%Different from the state-of-the-art approaches, 
We propose to utilize semantic experience, pattern visualization and temporal cycle transformation to understand action anticipation. 
% \textbf{SE}: Semantic Experience, \textbf{PV}: Pattern Visualization, \textbf{CC}: Cycle Consistency.
}

\centering
\resizebox{0.96\columnwidth}{!}{%
\renewcommand{\arraystretch}{1.4}
\begin{tabular}{M{0.25\columnwidth}|M{0.16\columnwidth}|M{0.17\columnwidth}|M{0.16\columnwidth}}
\toprule[1.2pt]
\multicolumn{1}{c|}{\multirow{3}{*}{Methods}} & \multicolumn{3}{c}{Settings}\\
\cline{2-4}

               & Semantic Experience &   Pattern Visualization  & Cycle Consistency \\ 
              
\midrule

2SCNN~\cite{simonyan2014two} & \textcolor{ao(english)}{\boldcheckmark} & \textcolor{cadmiumred}{\xmark}                   & \textcolor{cadmiumred}{\xmark} \\
\hline
RULSTM~\cite{furnari2019would} & \textcolor{cadmiumred}{\xmark}          & \textcolor{ao(english)}{\boldcheckmark}          & \textcolor{cadmiumred}{\xmark} \\
\hline
~\cite{liu2019forecasting} & \textcolor{ao(english)}{\boldcheckmark} & \textcolor{ao(english)}{\boldcheckmark}          & \textcolor{cadmiumred}{\xmark} \\
\hline

A-ACT (Ours)   & \textcolor{ao(english)}{\boldcheckmark} & \textcolor{ao(english)}{\boldcheckmark}          & \textcolor{ao(english)}{\boldcheckmark} \\

\bottomrule[1.2pt]

\end{tabular}

}
\label{tab:compare_methods}

\end{table}

\noindent\textbf{Early Action Recognition}. The goal of early action recognition is to recognize an ongoing action as early as possible given a partial observation of this action in an untrimmed video segment~\cite{de2016online}. It is important to model the sequential nature of the human activities for early action recognition. Earlier works formulate the task as a probabilistic framework and model the sequential nature of human activities using histograms of spatio-temporal features~\cite{ryoo2011human} and sparse-coding to estimate the likelihood of any activity ~\cite{cao2013recognize}. Some works propose variants of maximum-margin framework for training temporal event detectors for early action detection~\cite{hoai2014max, huang2014sequential}. Recently, Long-Short Term Memory (LSTM) networks are leveraged for this task due their powerful capability to model sequential data~\cite{sadegh2017encouraging, becattini2017done, de2018modeling, furnari2019would, shi2018action}. As opposed to early action recognition, we anticipate future action without any partial observations.

%However, the LSTMs network works sequentially and may not able able to focus on relevant input frames. 
\medskip
\noindent\textbf{Action Anticipation.} In action anticipation, the aim is to forecast the action that will happen in future. Unlike early action recognition, in anticipation we do not observe any snippet of the future action. Recently, human action anticipation in egocentric view has garnered a lot of interest~\cite{furnari2019would, qi2021self}. A simple strategy for action anticipation is to learn a direct mapping between observed visual cues and the  future  action  using  supervised  methods~\cite{simonyan2014two, wang2016temporal}. However, learning a direct mapping between distant time steps by only utilizing semantic past information can be challenging due to the weak correlation between the time steps as demonstrated in ~\cite{furnari2019would}. Hand-object contact information is used in~\cite{dessalene2021forecasting} utilizing contact anticipation maps and next-active object segmentation to learn features for action anticipation. Authors in~\cite{vondrick2016anticipating, furnari2019would} performs a regression based self-supervised pre-training of the LSTM network by predicting future representations to incorporate future information and then finetune the model for action anticipation. Reinforced Encoder-Decoder (RED) network is proposed in~\cite{gao2017red} to provide sequence-level supervision using inverse reinforcement learning. 
%All the observed frames may not be important for action anticipation. 
Methods with LSTM networks represent all the information from input frames in a single context vector and may not be able to focus on frames more relevant for anticipation. In contrast to these methods, we propose to jointly predict the representations and action labels to exploit the high correlation between them and utilize self-attention to give more emphasis on important features of the observed frames.\medskip

% Although authors in ~\cite{furnari2019would} employ attention to fuse different modalities, they use a single context vector to represent the input sequence, which might lead to loss of information as show.\medskip

\begin{figure*}[t]
% \vspace{2mm}
        \centering
        \includegraphics[width=0.98\textwidth]{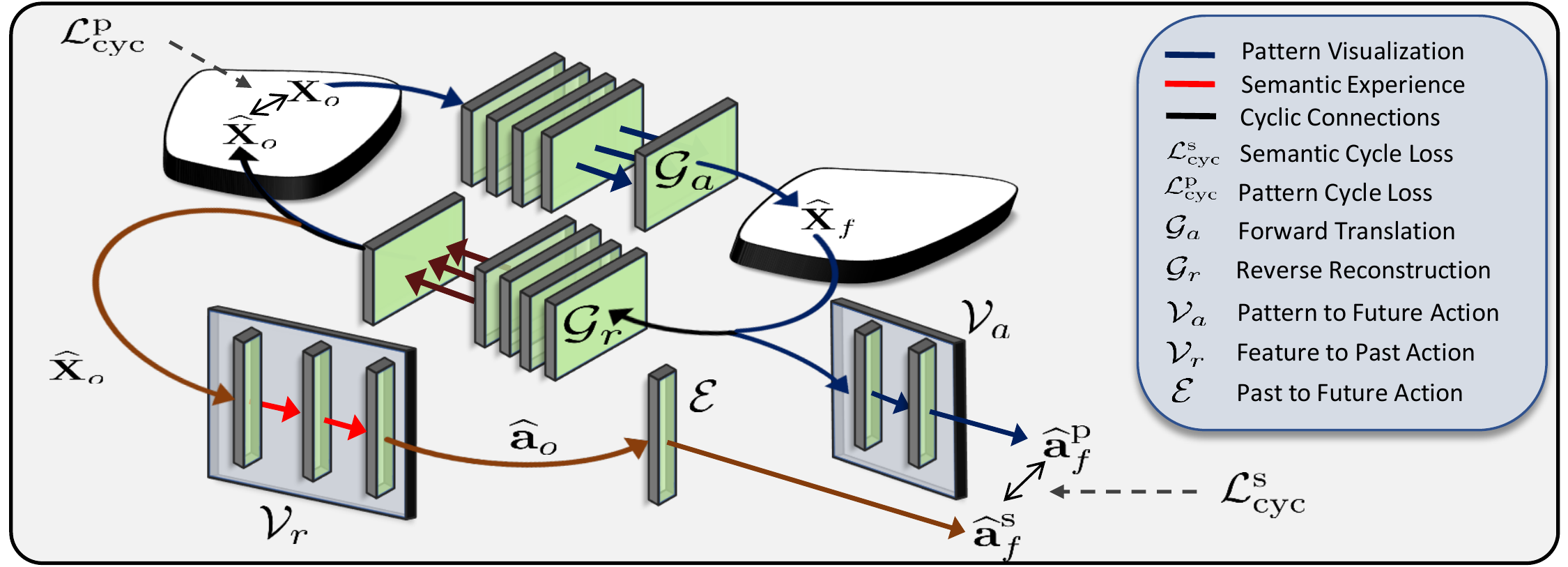}
        % \vspace{2mm}
        \caption{\textbf{Overview of the proposed cycle-transformation model.} Temporal cycle-transformation is enforced in the semantic label space by minimizing the distribution between $\mathbf{\widehat{a}}_f^{~s}$ and $\mathbf{\widehat{a}}_f^{~p}$, and in feature space by minimizing the reconstruction error between {$\mathbf{\widehat{X}}_o$} and {$\mathbf{X}_o$}. Additionally, mean-squared error loss between visualized feature  $\mathbf{\widehat{X}}_f$ and ground-truth features {$\mathbf{X}_f$}, cross-entropy loss between inferred action  {$\mathbf{\widehat{a}}_o$} and observed action label $\mathbf{a}_o$ and cross-entropy-loss between {$\mathbf{\widehat{a}}_f^{~p}$} and $\mathbf{a}_f$ are also applied while training the proposed model. Refer section~\ref{sec:cyc} for details.
        %The cycle-consistency in semantic space provides reasoning for the future action and the feature cycle-consistency helps learn discrimination features. 
        % \amit{This is only showing the cycle consistency model. It does not show the past experience and future visualization models. Can we incorporate those, or have a separate figure? Maybe have a different high level figure with less details, and then this figure that has all three parts?} \akash{Anticipation and reasoning - use semantics and patterns, and add $\widehat{a}_f$}
        }
        % \vspace{-4mm}
        \label{fig:overview}
\end{figure*}

\noindent\textbf{Cycle Consistency.} Cycle consistency has shown exceptional performance in spatial domain for tasks like image-to-image translation~\cite{isola2017image, zhu2017unpaired,huang2018multimodal}, video alignment~\cite{dwibedi2019temporal, wang2019learning}, image matching~\cite{Zhou_2015_ICCV, wang2018multi} and segmentation~\cite{toldo2020unsupervised, Zhang_2018_CVPR, mondal2019revisiting}. Recently, some works explored the concept of temporal cycle consistency in videos~\cite{ wang2019learning, farha2020long}. An approach to learn representations using video alignment as a proxy task and cycle consistency for training is proposed in~\cite{dwibedi2019temporal}. Authors in~\cite{wang2016temporal} exploit consistency in appearance between consecutive video frames to learn discriminative representations. Long term anticipation of activities is studied in~\cite{farha2020long} using cycle consistency. However, they only enforce cycle consistency in label space. Also, attention with RNN model is used as the context feature. RNN models process one step at a time which can learn local correlation between features well as opposed to global correlation. Unlike these methods, our approach utilizes cycle consistency in the label space as well as the feature space (refer Figure~\ref{fig:overview}). We also employ self-attention module as in transformer models to exploit local, as well as, global correlation between features. Experiments show that the cycle consistency in our model performs better than others. %than that of consistency in~\cite{farha2020long}.

% ------------------ Methodology

\section{Methodology}

We propose a framework for \textbf{A}ction \textbf{A}nticipation using \textbf{C}ycle \textbf{T}ransformations (\textbf{A-ACT}).
Our goal is to integrate the two human anticipation mechanisms into an anticipation framework by incorporating  the semantic information as well as future possibility to improve the performance of action anticipation model. We also introduce cycle-transformation in semantic label and feature space, to instill the human capability of reasoning in our models for the task of anticipation~\cite{breska2018double}.

\subsection{Problem Formulation}
Given an observed video sequence associated with the action $\textbf{a}_o$, our goal is to predict the next action $\textbf{a}_f$ for a given time horizon. 
Let the feature representations for frames of the observed video sequence corresponding to the action $\textbf{a}_o$ be represented as 
$\textbf{X}_o = [~\mathsf{x}_1, ~\mathsf{x}_2,\cdots,~\mathsf{x}_M ]$
and the representations for future frames corresponding to $\textbf{a}_f$  as %
$\textbf{X}_f = [ ~\mathsf{x}_{M'}, ~\mathsf{x}_{M'+1}, \cdots,~\mathsf{x}_{M'+N}]$ such that $M' = M+k$, 
where $x_i \in\mathbb{R}^{d}$ is a d-dimensional feature representation of $i^{th}$ frame, $k$ is the time horizon, $M$ and $N$ are the number of frames in the observed and future sequence, respectively. Our objective is to anticipate action $\textbf{a}_f$ using different models that leverage the semantic experience and perform pattern visualization, and integrate these models through cycle-transformations. 
%We also aim to study how well the human ability to anticipate future can be transferred to machine learning algorithms.

\subsection{Action Anticipation Models}
In this section, we describe the two action anticipation models and our proposed framework in detail. \medskip
%The semantic experience model $\mathbf{SE}$ first recognizes the observed action and then tries to anticipate future action based on the observed action. On the other hand, the pattern visualization model $\mathbf{PV}$ anticipates the possible future features and then performs action anticipation. We propose a framework that utilizes the semantic experience as well as the pattern visualization model along with temporal cycle-consistency in feature and label space for the task of anticipation. % Cycle-transformation framework $\mathbf{C}$ enforces the temporal cycle-consistency in features as well label space.

% Let us define a model $\mathcal{G}_F$ that take the observed frames as input and generate 
\subsubsection{Semantic Experience Model}\label{sec:SE}
The semantic experience model $\mathbf{SE}$ consists of an action recognition module $\mathcal{G}_o$ and an action anticipation layer $\mathcal{E}$ (top-branch in Fig.~\ref{fig:se-pv}).
% \amit{I think this should be Fig. 3.}
%relies completely on the observed representations for the task of action anticipation. 
The action recognition module $\mathcal{G}_o$ is utilized to recognize the observed action $\widehat{\textbf{a}}_o$ first from the representations of the observed frames $\textbf{X}_o$ as represented by ~\eqref{eqn:xo2ao}. Then, using the observed action label, the anticipation layer $\mathcal{E}$ anticipates the future action $\textbf{a}_f$ denoted by ~\eqref{eqn:ao2af}.
\begin{alignat}{3}
    \widehat{\textbf{a}}_o 
    &= \mathcal{G}_o\Big{(} [ ~\mathsf{x}_1, ~\mathsf{x}_2, \cdots, ~\mathsf{x}_M ]\Big{)} 
    \label{eqn:xo2ao}\\
    \widehat{\textbf{a}}_f^{~s}
    &= \mathcal{E}\big{(}~\widehat{\textbf{a}}_o\big{)}
    \label{eqn:ao2af}
\end{alignat}
 
\noindent where, $\widehat{\textbf{a}}_o$ and $\widehat{\textbf{a}}_f^{~s}$ are the inferred observed action labels and predicted future action labels, respectively. Since we first recognize the observed action, the anticipation layer relies completely on the past observation for future anticipation.\medskip

\noindent \textbf{Objective Function.} The objective function for the semantic past experience model consists of classification loss for the past action and the future action. It is defined as:
\begin{align}
    \mathcal{L}_\mathbf{S} = \mathcal{L}\Big(\widehat{\textbf{a}}_o,~ \textbf{a}_o\Big) + \mathcal{L}\Big(\widehat{\textbf{a}}_f^{~s},~ \textbf{a}_f\Big)
    \label{eqn:sem}
\end{align}
\noindent where, $\mathcal{L}$ is the categorical cross-entropy loss between predicted action labels and ground truth action labels.\bigskip

\subsubsection{Pattern Visualization Model}\label{sec:PV}
Given the feature representations of the observed frames, the pattern visualization module $\mathbf{PV}$ synthesizes possible future representations and then performs anticipation on the generated features. This is achieved by utilizing a feature translation module $\mathcal{G}_t$ which translates the features of observed frames $\textbf{X}_o$ into the features of future frames 
$\widehat{\textbf{X}}_f= [ \widehat{~\mathsf{x}}_{M'}, ~\widehat{\mathsf{x}}_{M'+1}, \cdots,~\widehat{\mathsf{x}}_{M'+N}]$ 
and then performs the action recognition through model $\mathcal{V}$ on the generated features to anticipate the action $\widehat{\textbf{a}}_f^{~p}$. The feature translation step and the action anticipation step for this model are given by the equations~\eqref{eqn:xo2xf} and~\eqref{eqn:xf2af} below.
\begin{alignat}{3}
    \widehat{\textbf{X}}_f 
    &= \mathcal{G}_t\Big{(} [ ~\mathsf{x}_1, ~\mathsf{x}_2, \cdots, ~\mathsf{x}_M ]\Big{)} \label{eqn:xo2xf}\\
    \widehat{\textbf{a}}_f^{~p}
    &= \mathcal{V}\big{(}~\widehat{\textbf{X}}_f\big{)}
    \label{eqn:xf2af}
\end{alignat}

\medskip
\noindent \textbf{Objective Function.} The objective function for pattern visualization model is defined by~\eqref{eqn:vis}.  It is comprised of a reconstruction loss between the generated future features and the ground truth future features, available during training of the model and an action anticipation loss for semantic action label generated for future action.
\begin{align}
    \mathcal{L}_\mathbf{P} = \mathcal{L}\Big(\widehat{\textbf{X}}_f,~ \textbf{X}_f\Big) + \mathcal{L}\Big(\widehat{\textbf{a}}_f^{~p},~ \textbf{a}_f\Big)
    \label{eqn:vis}
\end{align}
\noindent where, $\mathcal{L}\Big(\widehat{\textbf{X}}_f,~ \textbf{X}_f\Big)$ is the mean-squared error loss between the ground truth and generated features and $\mathcal{L}$ is the categorical cross-entropy loss for anticipated action.\medskip

% \noindent where, $N$ is the number of generated frames and $M'$ is the start of the action $M' = M + k$,  are the 
\subsubsection{Cycle Transformation Model}\label{sec:cyc}
 %Humans have ability to reason about the possible set of actions that leads to a particular outcome. This is because once you know the future the certainty of what happened in the past for a set of action increases. Therefore, 
 Anticipation of the future from the observed sequence should be consistent with the reasoning of the past given the anticipated future. We propose to incorporate this ability using the cycle consistency in feature as well as label space for our model.
 
 The cycle consistency model is composed of two feature translation modules ($\mathcal{G}_{a}$, $\mathcal{G}_{r}$) and two recognition modules ($\mathcal{V}_r$, $\mathcal{V}_a$) for past action and future action. And experience model $\mathcal{E}$ for anticipation action label using semantic past label.
 Cycle-transformation in temporal dimension is applied between the feature translation modules as well as output of the recognition modules.
 Given the features  $\textbf{X}_o$ of an observed video sequence, the forward translation module $\mathcal{G}_{a}$ translates the features in observed action space to the features in future action space $\widehat{\textbf{X}}_f$ given by~\eqref{eqn:frwd}. 
 Then the future recognition module $\mathcal{V}_a$ predicts the future action label $\widehat{\textbf{a}}_f^{~p}$ from features $\widehat{\textbf{X}}_f$ as represented by~\eqref{eqn:ant}.
 Next the reverse translation module $\mathcal{G}_{r}$ utilizes predicted features $\widehat{\textbf{X}}_f$ to reconstruct the observed $\widehat{\textbf{X}}_o$ using~\eqref{eqn:back}.
 The feature cycle-consistency is applied between the input observed features and the reconstructed observed features as illustrated in Figure~\ref{fig:overview}.
 Since the reconstructed features should be related to the observed video, these features are used to recognize the past action $\textbf{a}_o$ using the action recognition module $\mathcal{V}_r$. Then the semantic experience module $\mathcal{E}$ utilizes the semantic past labels, inferred using module $\mathcal{V}_r$, to anticipate future $\textbf{a}_f^s$ as shown below. %refer~\eqref{eqn:recog}.
%  \fontsize{9.5}{9.5} \selectfont{
\begin{subequations}
 \begin{alignat}{3}
 \centering
    & \text{Pattern Visualization} &: &&\quad \widehat{\textbf{X}}_f ~
    &= \mathcal{G}_{a}\Big{(} \textbf{X}_o 
    % [ ~\mathsf{x}_1, ~\mathsf{x}_2, \cdots, ~\mathsf{x}_M ]
    \Big{)}
    \label{eqn:frwd}\\
    &\text{Action Anticipation} &: &&\quad \widehat{\textbf{a}}_f^{~p} 
    &= \mathcal{V}_a\Big{(}\widehat{\textbf{X}}_f
    %[ ~\mathsf{x}_1, ~\mathsf{x}_2, \cdots, ~\mathsf{x}_M ]
    \Big{)}
    \label{eqn:ant}\\
    &\text{Feature Reconstruction} &: &&\quad \widehat{\textbf{X}}_o ~
    &= \mathcal{G}_{r}\Big{(}\widehat{\textbf{X}}_f
    %[ ~\mathsf{x}_{M'}, ~\mathsf{x}_{M'+1}, \cdots, ~\mathsf{x}_{M' + N} ]
    \Big{)} 
    \label{eqn:back}\\
    &\text{Semantic Recognition} &: &&\quad \widehat{\textbf{a}}_o ~
    &= \mathcal{V}_r\Big{(}\widehat{\textbf{X}}_o
    %[ ~\mathsf{x}_1, ~\mathsf{x}_2, \cdots, ~\mathsf{x}_M ]
    \Big{)}
    \label{eqn:recog}\\
    &\text{Semantic Anticipation} &: &&\quad \widehat{\textbf{a}}_f^{~s}
    &= \mathcal{E}\Big{(}\widehat{\textbf{a}}_o
    %[ ~\mathsf{x}_1, ~\mathsf{x}_2, \cdots, ~\mathsf{x}_M ]
    \Big{)}
    \label{eqn:exp}
\end{alignat}
 
\end{subequations}
% }
% \normalsize
% \smallskip

\noindent \textbf{Objective Function.} The cycle-consistency loss $\mathcal{L}_{\mathbf{C}}$ is imposed by minimizing the $\ell_2$ distance between the observed features $\textbf{X}_o$ and reconstructed observed features $\widehat{\textbf{X}}_o$ as cycle loss $\mathcal{L}_{cyc}^{~p}$ in feature space and the entropy loss between semantic anticipated action $\widehat{\textbf{a}}_f^{s}$ and pattern anticipated action $\widehat{\textbf{a}}_f^{~p}$ (ground truth) as cycle loss in semantic label space such that
\begin{subequations}
%  \begin{alignat}{3}
%  \centering
\begin{tabularx}{\hsize}{@{}XXX@{}}
     \begin{equation}
        \resizebox{0.325\columnwidth}{!}{%
        $\mathcal{L}_{cyc}^{~p} = \mathcal{L}\Big{(}\widehat{\textbf{X}}_o,\textbf{X}_o \Big{)}$
        }
    \label{eqn:cyc_p}
\end{equation} &
\begin{equation}
    \resizebox{0.325\columnwidth}{!}{%
        $\mathcal{L}_{cyc}^{~s} =
    \mathcal{L}\Big{(}\widehat{\textbf{a}}_f^{s},\widehat{\textbf{a}}_f^{~p} \Big{)}$
    }
    \label{eqn:cyc_s}
 \end{equation}
   \end{tabularx}
% \end{alignat}
\end{subequations}
% \vspace{-2mm}
\begin{align}
    \mathcal{L}_{\mathbf{C}} = \mathcal{L}_{cyc}^{~p} + \mathcal{L}_{cyc}^{~s}
    \label{eqn:cyc}
\end{align}
% \begin{subequations}
%     \begin{tabularx}{1.3\columnwidth}{Xp{-0.5cm}X}
%         \begin{equation}
%         \mathcal{L}_{cyc}^{~p} = \mathcal{L}_{mse}\Big{(}\widehat{\textbf{X}}_o,~\textbf{X}_o \Big{)}
%         \label{eqn:cyc_p}
%         \end{equation}
%     & &
%         \begin{equation}
%         \mathcal{L}_{cyc}^{~s} = \mathcal{L}\Big{(}\widehat{\textbf{a}}_f^{s},~\widehat{\textbf{a}}_f^{p} \Big{)}
%     \label{eqn:cyc_s}
%         \end{equation}
% \end{tabularx}
% \end{subequations}
\noindent where $\mathcal{L}\Big{(}\widehat{\textbf{X}}_o,\textbf{X}_o \Big{)}$ is the mean-squared error loss between the ground truth observed features and reconstructed features and $\mathcal{L}\Big{(}\widehat{\textbf{a}}_f^{~s},\widehat{\textbf{a}}_f^{~p} \Big{)}$ is the categorical cross-entropy loss for anticipated action.
\subsection{Overall Objective Function}
The overall objective function is composed of the semantic experience loss $\mathcal{L}_{\mathbf{S}}$ as defined in~\eqref{eqn:sem}, the pattern visualization loss $\mathcal{L}_{\mathbf{P}}$ as in~\eqref{eqn:vis} and the cycle-consistency loss. 
\begin{align}
    \mathcal{L} = \lambda_s~ \mathcal{L}_{\mathbf{S}} + \lambda_p~\mathcal{L}_{\mathbf{P}} + \lambda_{c}~\mathcal{L}_{\mathbf{C}} 
    \label{eqn:C}
\end{align}
\noindent where, $\lambda_s, \lambda_p,$ and $\lambda_c$ are the regularization constants for semantic experience loss, pattern visualization loss and cycle-consistency loss, respectively.

% \subsubsection{Self-Attention}
% \akash{Will add details here}

\subsection{Network Architecture}

We choose a two layer self-attention modified transformer model~\cite{vaswani2017attention} with 8 multi-head attention as the backbone architecture for the action recognition module $\mathcal{G}_o$ for observed sequence and the feature translation modules $\mathcal{G}_t, \mathcal{G}_a$ and  $\mathcal{G}_r$. The input dimension of the self-attention model is same as the observed feature dimension and the hidden dimension is half of input dimension. The semantic recognition layer $\mathcal{V}_r$, the experience layer $\mathcal{E}$ and the visualization layers $\mathcal{V}_a$ are two layer multi-layer perceptrons with input, output and hidden dimension same as the dimension of feature/label of observed video frames. 

% We use Adam with a learning rate $5 \times 10^{-5}$ and weight decay $0.01$ to train all our models. The loss regularization constants $\lambda_r$ and $\lambda_v$ are set to 1 and $\lambda_{cyc}$ is set 0.1 for all experiments. All the models are trained with a batch size of 128.

% ------------------ Experiments
% \newpage
\section{Experiments}

We perform rigorous experiments on egocentric action anticipation dataset and the procedural-activities datasets to study the impact of different action anticipation models. The datasets are discussed below.\medskip

% In this section, we first introduce the datasets along with their evaluation protocol and then provide the implementation details of our approach.

% \input{tables/2_different_timesteps}

% \subsection{Datasets}
%We evaluate the performance of different anticipation models on publicly available EPIC-Kitchens 55 dataset~\cite{damen2018scaling}, Breakfast dataset~\cite{kuehne2014language} and 50Salad dataset.

\noindent \textbf{EPIC-Kitchens 55.} 
The EPIC-Kitchens 55~\cite{damen2018scaling} dataset is a collection of 55 hours of videos with overall 11.4M frames. All the videos comprise of the daily kitchen activities in egocentric view. It provides 2513 fine-grained action annotations along with verb and noun annotation for each segment containing any action. The dataset is divided into training, validation and two test sets (seen and unseen). The seen test set consists of the kitchen environment seen in the training dataset whereas the unseen test set contains new kitchen environment to evaluate the generalizability of any algorithm in unseen environment. We evaluate different anticipation models on the validation set. We use the validation set for hyper-parameter search and compare the results of our proposed approach with other state-of-the-art methods on both the test sets~\cite{furnari2019would}.\medskip

\noindent \textbf{Procedural Activities.} The procedural activities datasets consists of Breakfast dataset which contains videos of cooking activities for preparation of 10 dishes and Salad50 dataset which consists of videos of people mixing 2 salads each. These datasets are discussed below.\smallskip

\textbf{Breakfast dataset.} The Breakfast dataset is a large-scale 1,712 videos, with a total duration of ~66.7 hours, where each video belongs to one out of ten breakfast related activities in third-person view~\cite{kuehne2014language}. The video frames are annotated with 48 coarser actions specific to breakfast related activities. On average, each video contains 6 action instances and is 2.3 minutes long. % \noindent \textbf{Features.}\medski
%\textbf{Evaluation Protocol.} 
Evaluation of different anticipation models is performed by taking average of performance measure over the 4 standard splits as proposed in~\cite{kuehne2016end}.\smallskip

\textbf{50Salads dataset.} 
The 50Salads dataset~\cite{stein2013combining} contains videos of people preparing different kinds of salad. There are 50 videos with average duration 6.4 minutes and contain 20 action instances per video. It features 17 fine-grained action labels like cut tomato or peel cucumber.% \noindent \textbf{Features.}\medskip 
%\textbf{Evaluation Protocol.}
We follow the evaluation protocol propose in~\cite{stein2013combining} by performing a five-fold cross-validation and report the average performance.

\begin{table}[t]
    \caption{\textbf{EPIC-Kitchen dataset.}Performance of different anticipation models at 1s before the action start time. Our model with cycle transformation outperform other anticipation models by +2.6\% in Top-1 accuracy and +3.5\% in Top-5 accuracy when using appearance features.}
    \label{tab:appearance_feat}
    \centering
    \resizebox{0.96\columnwidth}{!}{%
    \renewcommand{\arraystretch}{1.3}
    \begin{tabular}{>{\centering\arraybackslash}p{2.5cm}|>{\centering\arraybackslash}p{2cm}|>{\centering\arraybackslash}p{2cm}}

        \toprule[1.2pt]
        \multirow{2}{*}{Model} & \multicolumn{2}{c}{\textbf{Anticipation @1s}} \\ \cline{2-3} 
                               & Top-1 (\%)        & Top-5 (\%)        \\ 
                               \toprule[1.2pt]
        $\mathbf{SE}$                      & 12.11                 & 31.35                 \\ \hline
        $\mathbf{PV}$                       & 12.54                  & 32.70                   \\ \hline
        \textbf{A-ACT} (ours)                     & 14.70                   & 34.83     \\
        \toprule[1.2pt]
    
    \end{tabular}
    }
\end{table}
% \input{plots/fig_ek_1s}
% \input{tables/2_different_timesteps}

% ------------------ Results
\subsection{Action Anticipation on EPIC-Kitchens}
\begin{figure}[!t]
\resizebox{0.97\columnwidth}{!}{
\pgfkeys{
   /pgf/number format/.cd, 
    %   set decimal separator={.{\!}},
      set thousands separator={}
}
\pgfplotsset{
   every axis/.append style = {
      line width = 1pt,
      tick style = {line width=1pt}
   }
}

\begin{tikzpicture}
% [X/.style = {circle, fill=black, inner sep=1.5pt, 
            % label={[font=\scriptsize]below right:#1},
            % node contents={}}]

\begin{axis}[
	xlabel={Anticipation Time (sec) before the action},
	ylabel={Top-5 Anticipation Accuracy (\%)},
	xmin=0.25, xmax=2,
	ymin=24.1, ymax=38,
	legend pos=north west,
    ymajorgrids=true,
    xmajorgrids=true,
    grid style=dashed,
    x dir=reverse,
    xtick={0, 0.25, 0.5, 0.75, 1, 1.25, 1.5, 1.75, 2},
]
% \begin{axis}[
% 	x tick label style={
% 		/pgf/number format/1000 sep=},
% 	ylabel=Year,
% 	enlargelimits=0.05,
% 	legend style={at={(0.5,-0.1)},
% 	anchor=north,legend columns=-1},
% 	ybar interval=0.1,
% ]
\addplot
	coordinates {(2,  26.70) (1.75, 28.03) (1.5, 29.10) (1.25, 30.17) (1, 31.35) (0.75, 32.54) (0.5, 33.66) (0.25, 35.01)};
\addplot
	coordinates {(2,  26.35) (1.75, 28.15) (1.5, 29.71) (1.25, 30.99) (1, 32.70) (0.75, 33.18) (0.5, 34.06) (0.25, 35.30)};
\addplot+[brown, mark=triangle*, mark size=3.25pt]
    coordinates {(2,  26.92) (1.75, 28.40) (1.5, 30.31) (1.25, 32.35) (1, 34.83) (0.75, 35.03) (0.5, 35.70 ) (0.25, 36.52)};
    
    % \draw[dashed] (0,0) -- (0, 38) node[X={$N=1$}] -- (0,-0.2);

\legend{SE, PV, A-ACT}
\end{axis}
\end{tikzpicture}
}
\caption{\textbf{Anticipation robustness of different anticipation models.} The cycle-transformation model \textbf{A-ACT} performs significantly better for time steps 1.25s, 1s and 0.75s as compared to other time steps. The performance of the visualization model \textbf{PV} degrades as the time horizon increase which can be seen for anticipation at 2s.}
\label{fig:robustness}
\end{figure}
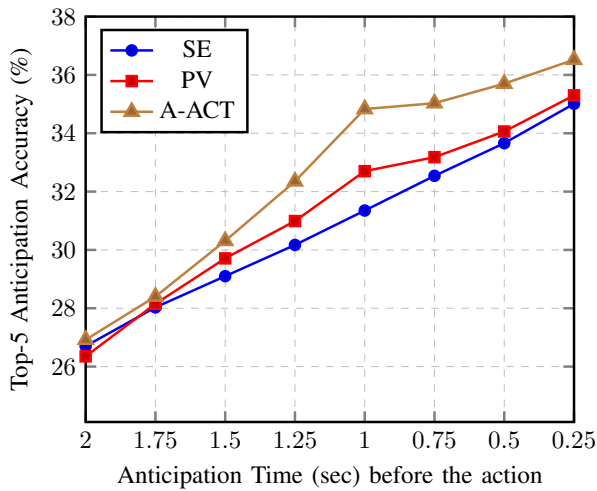
We study the impact of different anticipation models on the task of action anticipation in the EPIC-Kitchens dataset. We first study the standard anticipation task at 1s before the action start time using different anticipation models. Then we present ablation analysis on anticipation at different time steps to study the robustness of the anticipation models. We also conduct ablation analysis for the impact of different cycle-transformations, loss components on the performance and efficiency of different models in a setting with limited data.\medskip

\begin{table}[t]
    \caption{\textbf{EPIC-Kitchen: Impact of different cycle transformations.} It can be observed that cycle in feature space performs better than cycle in semantic space. Also, the performance improvement when using both the cycle-transformation suggests that the cycle in the label space and  the feature space is complimentary to each other.}
    \label{tab:ablation_cycle}
    \centering
    \resizebox{0.99\columnwidth}{!}{%
    \renewcommand{\arraystretch}{1.4}
\begin{tabular}{>{\arraybackslash}p{2.6cm}|>{\centering\arraybackslash}p{2cm}|>{\centering\arraybackslash}p{2cm}}
\toprule[1.2pt]
\multirow{2}{*}{\textbf{Cycle Transform}} & \multicolumn{2}{c}{\textbf{Anticipation @1s}} \\ \cline{2-3} 
                       & Top-1 (\%)        & Top-5 (\%)        \\ 
                       \toprule[1.2pt]
Semantic                      & 12.24                 & 31.80                 \\ \hline
Feature                      & 12.63                  & 33.07                   \\ \hline
Both                     & 14.70                   & 34.83     \\
\toprule[1.2pt]
\end{tabular}
}
\end{table}
\begin{table}[t]
    \caption{\textbf{EPIC-Kitchen: Impact of loss terms on the validation set.} Using both the cycle-transformation we achieve a performance boost of about 1.2\% in top-1 accuracy and 2.2\% in top-5\% accuracy.}
    \label{tab:ablation_loss_ep}
    \centering
    \resizebox{0.97\columnwidth}{!}{%
    \renewcommand{\arraystretch}{1.7}
\begin{tabular}{l|c|c}
% {>{\raggedright\arraybackslash}p{4.6cm}|>{\centering\arraybackslash}p{1.6cm}|>{\centering\arraybackslash}p{1.6cm}}
\toprule[1.2pt]
\multirow{2}{*}{\textbf{Loss Terms}} & \multicolumn{2}{c}{\textbf{Anticipation @1s}} \\ \cline{2-3} 
                       & Top-1 (\%)        & Top-5 (\%)        \\ 
                       \toprule[1.2pt]

$ \mathcal{L}_{cyc}^{p} + \mathcal{L}\big(\widehat{\textbf{a}}_f^{~p},~ \textbf{a}_f\big)$         
& 13.22                 & 32.19                 \\ 
% + \mathcal{L}_c(\hat{a}_f, a_f)
\hline
$\mathcal{L}_{cyc}^{p} +  \mathcal{L}\big(\widehat{\textbf{a}}_f^{~p},~ \textbf{a}_f\big)                   + \mathcal{L}\big(\widehat{\textbf{a}}_o,~ \textbf{a}_o\big)$  
& 13.47                  & 32.67\\ \hline
$ \mathcal{L}_{cyc}^{p} + \mathcal{L}_{cyc}^{s} + \mathcal{L}_{\textbf{S}} +  \mathcal{L}_{\textbf{P}}$                     & 14.70                   & 34.83\\
\toprule[1.2pt]

\end{tabular}
}

\end{table}
\begin{table}[!tbh]
% \vspace{1mm}
\caption{\textbf{EPIC Kitchen: Impact of training dataset size.} Our approach \textbf{A-ACT} works reasonably well even when using 50\% of the data.}

\label{tab:semi}
\centering
\resizebox{0.97\columnwidth}{!}{%
\renewcommand{\arraystretch}{1.4}
\begin{tabular}{c|c|c|c|c|c}
\toprule[1.2pt]
\textbf{Cycle} & \textbf{Pretrained} & \textbf{10\%} & \textbf{20\%} & \textbf{30\%} & \textbf{50\%} \\
\toprule[1.2pt]
%16.59 23.12 25.86 27.44
\textcolor{cadmiumred}{\xmark} &  \textcolor{cadmiumred}{\xmark}      & 15.76     & 22.83     & 25.11     & 26.84   \\ \hline
\textcolor{ao(english)}{\boldcheckmark}     &   \textcolor{cadmiumred}{\xmark}   & 17.23     & 24.37     & 26.16     & 28.02   \\ \hline
\textcolor{ao(english)}{\boldcheckmark}  & \textcolor{ao(english)}{\boldcheckmark} & 18.90     & 26.14     & 28.20     & 29.71   \\
\toprule[1.2pt]

\end{tabular}
}

\end{table}

\begin{table*}[!tbh]
% \vspace{1.5em}
\caption{\textbf{EPIC-Kitchen: Comparison with the state-of-the-art methods on the test dataset.} We evaluate a late feature fusion cycle-transformation model with the state-of-the-art methods. It can be observed that the proposed approach \textbf{A-ACT} not only outperforms the state-of-the-art methods using pre-computed features~\cite{furnari2019would} but is very competitive against method using superior features (Action-Bank~\cite{wu2019long}) and additional supervised features (Ego-OMG~\cite{ dessalene2020egocentric}).}
\label{tab:test_results}
\centering
\resizebox{0.985\linewidth}{!}{%
\renewcommand{\arraystretch}{1.46}
\begin{tabular}{c@{\hspace{1em}}l|ccc|ccc|ccc|ccc}
\toprule[1.2pt]
    & \multirow{2}{*}{\textbf{Methods}} & \multicolumn{3}{c|}{\textbf{Top-1 Accuracy}} & \multicolumn{3}{c|}{\textbf{Top-5 Accuracy}} & \multicolumn{3}{c|}{\textbf{Avg Class Precision}} & \multicolumn{3}{c}{\textbf{Avg Class Recall}} \\
    % \cmidrule(lr){3-5} \cmidrule(lr){6-14} \cmidrule(lr){9-11} \cmidrule(lr){12-14}
    \Cline{0.5pt}{3-14}
    &       & Verb     &   Noun    &  Action    & Verb     &   Noun    &  Action    & Verb     &   Noun    &  Action    & Verb     &   Noun    &  Action    \\
\toprule[1.2pt]
\parbox[t]{2.3mm}{\multirow{9}{*}{\rotatebox[origin=c]{90}{S1}}}
    &   DMR~\cite{wu2019long}          & 26.53 & 10.43 & 01.27  & 73.3  & 28.86 & 07.17  & 06.13  & 04.67  & 00.33 & 05.22  & 05.59  & 00.47 \\
    &   2SCNN        & 29.76 & 15.15 & 04.32  & 76.03 & 38.56 & 15.21 & 13.76 & 17.19 & 02.48 & 07.32  & 10.72 & 01.81 \\
    &   ATSN         & 31.81 & 16.22 & 06.00     & 76.56 & 42.15 & 28.21 & 23.91 & 19.13 & 03.13 & 09.33  & 11.93 & 02.39 \\
    &   Miech et al.~\cite{miech2019leveraging} & 30.74 & 16.47 & 09.74  & 76.21 & 42.72 & 25.44 & 12.42 & 16.67 & 03.67 & 08.80   & 12.66 & 03.85 \\
    \Cline{0.5pt}{2-14}
    &   RU-LSTM~\cite{furnari2019would}      & 33.04 & 22.78 & 14.39 &  79.55 & 50.95 &  33.73 &  25.50  & 24.12 &  07.37 &  15.73 &  19.81 &  07.66 
    \\
    \Cline{0.5pt}{2-14}
    &   \textbf{A-ACT} (ours) & \textbf{35.99} & \textbf{24.31} & \textbf{16.64} &  \textbf{80.12}     &   \textbf{53.52}    &  \textbf{36.73}  & \textbf{28.09}      & \textbf{25.25}     & \textbf{08.32}      & \textbf{18.36}      &   \textbf{20.80} & \textbf{8.45}   \\
    \Cline{0.5pt}{2-14}
    & Ego-OMG~\cite{ dessalene2020egocentric} & 32.20 & 24.90  & 16.02  & 77.42  & 50.24  & 34.53 & 14.92  & 23.25  & 04.03  & 15.48  & 19.16  & 05.36 \\
    &   Action Bank~\cite{wu2019long} & 37.87 & 24.10 & 16.64 & 79.74 & 53.98 & 36.06 & 36.41 & 25.20 & 09.64 & 15.67 & 22.01 & 10.05 \\
\midrule[0.1em]

\parbox[t]{2.3mm}{\multirow{9}{*}{\rotatebox[origin=c]{90}{S2}}} 
    &   DMR~\cite{wu2019long}          & 24.79 & 08.12 & 00.55 & 64.76 & 20.19 & 04.39 & 09.18 & 01.15 & 00.55 & 05.39 & 04.03  & 00.20 
    \\
    &   2SCNN       & 25.23 & 09.97 & 02.29 & 68.66 & 27.38 & 09.35 & 16.37 & 06.98 & 00.85 & 05.80 & 06.37  & 01.14 
    \\
    &   ATSN         & 25.3  & 10.41 & 02.39 & 68.32 & 29.50 & 06.63 & 07.63 & 08.79 & 00.80 & 06.06 & 06.74  & 01.07 
    \\
    &   Miech et al.~\cite{miech2019leveraging} & 28.37 & 12.43 & 07.24 & 69.96 & 32.20 & 19.29 & 11.62 & 08.36 & 02.20 & 07.80 & 09.94  & 03.36
    \\
    \Cline{0.5pt}{2-14}
    &   RU-LSTM~\cite{furnari2019would}      & 27.01 & 15.19 & 08.16 & 69.55 &34.38 & 21.10 & 13.69 & 09.87 & 03.64 & 09.21 & 11.97  &  04.83 
    \\
    &   IAI~\cite{zhang2020egocentric}          & 27.89 & 14.89 & 8.57 & 70.06 & 35.51 & 21.41 & -     & -     & -     & -     & -      & -
    \\
    % & LTCC & -     & -     & -     & -     & -      & - & -     & -     & -     & -     & -      & -
    % \\
    \Cline{0.5pt}{2-14}
    &   \textbf{A-ACT} (ours) &   \textbf{29.19}    &    \textbf{15.98}   &  \textbf{10.31}    & \textbf{71.12}       & \textbf{36.53}      & \textbf{23.46}     &     \textbf{16.21}  & \textbf{10.87}      &   \textbf{03.98}   &   \textbf{10.75}   &  \textbf{12.19}     & \textbf{5.33}     
    \\
    \Cline{0.5pt}{2-14}
    & Ego-OMG~\cite{ dessalene2020egocentric}   & 27.42 & 17.65 & 11.81 & 68.59 & 37.93 & 23.76 & 13.36 & 15.19 & 04.52 & 10.99 & 14.34 & 05.65 
    \\
    &   Action Bank~\cite{wu2019long}  & 29.50 & 16.52 & 10.04 & 70.13 & 37.83 & 23.42 & 20.43 & 12.95 & 04.92 & 08.03 & 12.84 & 06.26 
    \\
\toprule[1.2pt]
\end{tabular}
}

\end{table*}

\noindent \textbf{Anticipation @ 1s before the action.} We study the impact of different anticipation models on standard action anticipation task. The appearance features provided by authors in~\cite{furnari2019would} is used for this experiment. The performance of different anticipation models on the validation set is presented in Table~\ref{tab:appearance_feat}. We observe that our cycle-transformation model outperforms other anticipation models by a margin of $2.6\%$ in Top-1 accuracy and $3.0\%$ in Top-5 accuracy. It is interesting to note that the pattern visualization model $\textbf{PV}$ performs better than the semantic experience model $\textbf{SE}$. As the semantic space is likelihood of action based on given observed feature, we believe that the semantic experience model loses some details when the observed features are used to recognize the observed semantic label. Due to this the experience layer may not be able to anticipate future semantic well as opposed to pattern visualization layer. The pattern visualization layer first synthesis probable future features where details are preserved in the feature space. Thus action anticipation on the synthesized feature performs better than the semantic experience model.\medskip

\noindent \textbf{Anticipation Robustness.} We evaluate all the anticipation models for the task of anticipation at different time step. This is to evaluate the robustness of the anticipation models for near future as well further future.
%We perform experiments using all the three anticipation models to anticipate the action at different time steps. 
%This is to evaluate the robustness of the models for near future as well further future anticipation. 
Performance of the anticipation models at different time steps before the action is listed in Figure~\ref{fig:robustness}. It is evident that the cycle-transformation model $\textbf{A-ACT}$ outperforms the other models for all the time steps. The cycle-transformation model significantly outperforms the other models by a margin of $~2\%$ for time steps 1.25s, 1s and 0.75s. However, the improvement margin for very near future (left end of the graph) and further future (right end of the graph) is comparatively small. We believe the narrowing gap in the performance of different models, for further future, is due to the fact that the semantic experience $\textbf{SE}$ model and the pattern visualization $\textbf{PV}$ model now have access to more features and the temporal gap between the observed features and the future action is narrow. Hence, these models are very competitive to the cycle-transformations model.
% It can also be deduced that for the further future, it is the models learns equa. 
On the other hand, the uncertainty of the future feature generation increases with the time horizon. We conjecture that it is difficult for the pattern visualization model to capture very long-term correlation between the features. Hence, the semantic experience $\textbf{SE}$ model slightly outperforms or is at-par with the pattern visualization $\textbf{PV}$ model at anticipation times 2s and 1.75s. We conjecture that it is difficult for the pattern visualization model to capture very long-term correlation between the features.\medskip

\noindent \textbf{Impact of Cycle Transformations.} Here we investigate the importance of semantic label and feature cycle-transformation on the validation dataset using appearance features. We can observe from Table~\ref{tab:ablation_cycle} that the cycle in feature space performs better than cycle in label space in both Top-1 accuracy (12.24\% vs. 12.63\%) and Top-5 accuracy (31.80\% vs. 33.07\%), when compared for the anticipation task at 1s before the action. The performance is further improved to 14.70\% in Top-1 accuracy and to 34.83\% in Top-5 accuracy  when cycle-transformation is applied in the label as well as feature space. This suggests that the proposed cycle-transformation in the label and the feature space is able to learn discriminative features for the task of action anticipation.\medskip

\noindent \textbf{Impact of different loss terms}
The impact of different loss terms on anticipating the action on validation dataset at 1s before occurrence is presented in Table~\ref{tab:ablation_loss_ep}. We keep the cycle consistency loss $\mathcal{L}_{cyc}^{p}$ between the observed features and reconstructed observed feature consistent in the experiments to evaluate the impact of other loss terms. Since we are performing action anticipation a loss term  $\mathcal{L}\big(\widehat{\textbf{a}}_f^{~p},~ \textbf{a}_f\big)$ for anticipation is added to train the anticipation network.  We have the following observation from Table~\ref{tab:ablation_loss_ep}. (1) Action anticipation using only the pattern visualization loss $\mathcal{L}_{\textbf{P}}$ and consistency loss for observed feature $\mathcal{L}_{cyc}^{~p}$ achieves good performance in terms of Top-1 (13.22\%) and Top-5 accuracy (32.19\%). (2) Adding a recognition loss term $\mathcal{L}(\widehat{\textbf{a}}_o,~ \textbf{a}_o)$ for observed actions further improves the performance as additional supervision from observed action labels is provided. However, there is only a slight boost in performance. We believe it could be due the cycle-consistency in feature space which helps learn in rich features for anticipation. (3) When training the model with the total loss as in~\eqref{eqn:C} there is a gain of about 1.2\% in top-1 and 2.2\% in top-5 accuracy.\medskip

\noindent \textbf{Availability of training data.} We conduct a small experiment on Epic-Kitchen to evaluate the performance of our framework with limited data in semi-supervised setup. As humans do not need more samples to learn a task, with this experiment we test the learning capability of our approach.  We generate training splits with 10\%, 20\%, 30\% and 50\% of total data by sampling data points of each class proportional to the number of samples in that class. From table~\ref{tab:semi}, we observe that our approach achieves a top-5 performance of 29.71\% with only 50\% of the data as oppose to 34.83\% using 100\% of the data.\medskip %We observed that 

% \begin{figure}[!t]
% \resizebox{0.96\columnwidth}{!}{
% \input{plots/tikz_bf_baselines}
% }
% \caption{\textbf{Breakfast dataset: Performance of different anticipation mechanisms.} }
% \label{fig:procedural_baselines}
% \end{figure}

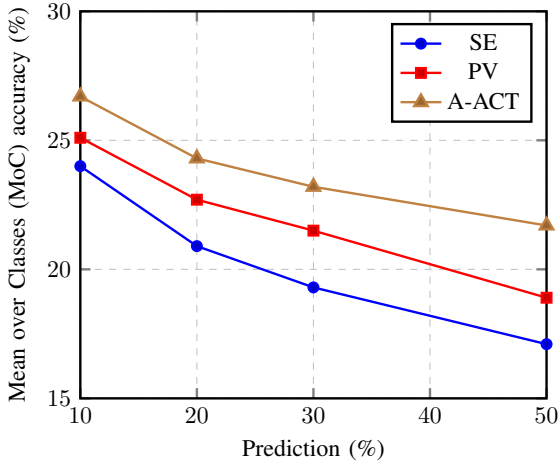
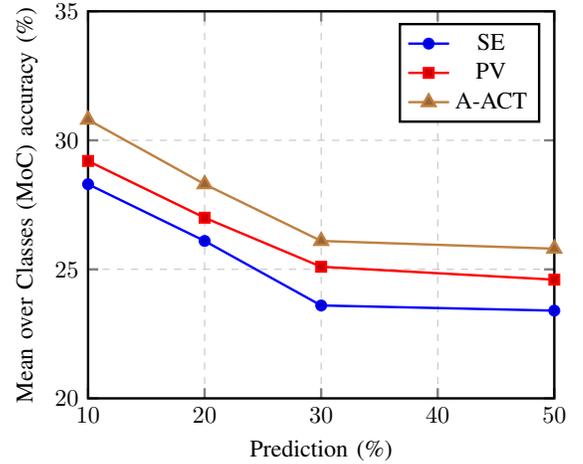
\begin{figure*}[!t]
    \captionsetup[sub]{font=large}
    %  \centering
     \begin{subfigure}[b]{0.46\textwidth}
        %  \centering
         \resizebox{0.955\textwidth}{!}{%
         \pgfkeys{
   /pgf/number format/.cd, 
    %   set decimal separator={.{\!}},
      set thousands separator={}
}
\pgfplotsset{
   every axis/.append style = {
      line width = 1pt,
      tick style = {line width=1pt}
   }
}

\begin{tikzpicture}

\begin{axis}[
    xlabel={Prediction (\%)},
	ylabel={ Mean over Classes (MoC) accuracy (\%)},
	xmin=10, xmax=50,
	ymin=15, ymax=30,
	legend pos=north east,
    ymajorgrids=true,
    xmajorgrids=true,
    grid style=dashed,
    xtick={10, 20, 30, 40, 50},
]
% \begin{axis}[
% 	x tick label style={
% 		/pgf/number format/1000 sep=},
% 	ylabel=Year,
% 	enlargelimits=0.05,
% 	legend style={at={(0.5,-0.1)},
% 	anchor=north,legend columns=-1},
% 	ybar interval=0.1,
% ]
\addplot
	coordinates {(10, 24.0) (20, 20.9) (30, 19.3) (50, 17.1)};
\addplot 
	coordinates {(10, 25.1) (20, 22.7) (30, 21.5) (50, 18.9)};
\addplot+[brown, mark=triangle*, mark size=3.25pt]
    coordinates {(10, 26.7) (20, 24.3) (30, 23.2) (50, 21.7)};
\legend{SE, PV, A-ACT}
\end{axis}
\end{tikzpicture}
         }
         \caption{Anticipation with 20\% observation.}
         \label{fig:procedural_baselines_a}
     \end{subfigure}
     \hfill
     \begin{subfigure}[b]{0.46\textwidth}
         \centering
         \resizebox{0.955\textwidth}{!}{%
         \pgfkeys{
   /pgf/number format/.cd, 
    %   set decimal separator={.{\!}},
      set thousands separator={}
}
\pgfplotsset{
   every axis/.append style = {
      line width = 1pt,
      tick style = {line width=1pt}
   }
}

\begin{tikzpicture}

\begin{axis}[
    xlabel={Prediction (\%)},
	ylabel={ Mean over Classes (MoC) accuracy (\%)},
	xmin=10, xmax=50,
	ymin=20, ymax=35,
	legend pos=north east,
    ymajorgrids=true,
    xmajorgrids=true,
    grid style=dashed,
    xtick={10, 20, 30, 40, 50},
]
% \begin{axis}[
% 	x tick label style={
% 		/pgf/number format/1000 sep=},
% 	ylabel=Year,
% 	enlargelimits=0.05,
% 	legend style={at={(0.5,-0.1)},
% 	anchor=north,legend columns=-1},
% 	ybar interval=0.1,
% ]
\addplot
	coordinates {(10, 28.3) (20, 26.1) (30, 23.6) (50, 23.4)};
\addplot 
	coordinates {(10, 29.2) (20, 27.0) (30, 25.1) (50, 24.6)};
\addplot+[brown, mark=triangle*, mark size=3.25pt]
    coordinates {(10, 30.8) (20, 28.3) (30, 26.1) (50, 25.8)};
\legend{SE, PV, A-ACT}
\end{axis}
\end{tikzpicture}
         }
         \caption{Anticipation with 30\% observation.}
         \label{fig:procedural_baselines_b}
     \end{subfigure}
     \caption{\textbf{Breakfast Dataset: Performance of different anticipation mechanisms.} With increase in anticipation horizon the performance of all models degrades as expected. \textbf{PV} model outperforms \textbf{SE} for different predictions by observing 20\% and 30\% of the video. Our proposed \textbf{A-ACT} outperforms \textbf{PV} with a large margin of about 3\% at anticipating at 50\% with 20\% observation against \textbf{PV}. For 30\% observation, our proposed approach shows improvement of 0.8\% when compared with \textbf{PV}.}
     \label{fig:procedural_baselines}
\end{figure*}
\begin{table*}[t]
% \vspace{1mm}
\captionsetup{width=.85\textwidth}
\caption{\textbf{Breakfast dataset: Performance of our models with different loss terms.} It can be observed that cycle consistency in the feature space (row 3, $\mathcal{L}_{cyc}^{s}$) improves the performance as compared to only using cycle consistency in the label space. Note: $\mathcal{L}_{\textbf{S}}$ and $\mathcal{L}_{\textbf{P}}$ consists of the action classification loss $\mathcal{L}\big(\widehat{\textbf{a}}_o,~ \textbf{a}_o\big)$ and action anticipation loss $\mathcal{L}\big(\widehat{\textbf{a}}_f^{~p},~ \textbf{a}_f\big)$ as in row 2.}
\label{tab:ablation_loss_bf}
\centering
% \resizebox{\columnwidth}{!}{%
\renewcommand{\arraystretch}{1.3}
% \begin{tabular}{
% % >{\centering\arraybackslash}p{0.5cm}
% >{\raggedright\arraybackslash}p{4.5cm}|>{\centering\arraybackslash}p{0.5cm}>{\centering\arraybackslash}p{0.5cm}>{\centering\arraybackslash}p{0.5cm}>{\centering\arraybackslash}p{0.5cm}|>{\centering\arraybackslash}p{0.5cm}>{\centering\arraybackslash}p{0.5cm}>{\centering\arraybackslash}p{0.5cm}>{\centering\arraybackslash}p{0.5cm}}
\begin{tabular}{l|cccc|cccc}
\toprule[1.2pt]
\multirow{2}{*}{\textbf{Loss Terms}} & \multicolumn{4}{c|}{\textbf{20\%}}                                                                & \multicolumn{4}{c}{\textbf{30\%}}                                                            \\
\cline{2-9}
& 10\%                      & 20\%                     & 30\%                     & 50\%                       & 10\%                      & 20\%                      & 30\%                      & 
                               50\% \\
\toprule[1.2pt]
% \multicolumn{9}{l}{Breakfast} \\ \hline

% & \textbf{E}                             & 24.0                      & 20.9                      & 19.3                      & 17.0                                               & 28.3                      & 26.1                      & 23.6                      & 23.4                        \\
% & \textbf{V}                             & 25.1                      & 22.7                      & 21.5                      & 18.9                                               & 29.2                      & 27.0                      & 25.1                      & 24.6                         \\ \cline{2-10}
$ \mathcal{L}_{cyc}^{p} + \mathcal{L}\big(\widehat{\textbf{a}}_f^{~p},~ \textbf{a}_f\big)$   & 25.3  & 23.2  & 21.4 & 19.2 & 28.9 &  26.2 & 24.1  &  21.9\\

$\mathcal{L}_{cyc}^{p} +  \mathcal{L}\big(\widehat{\textbf{a}}_f^{~p},~ \textbf{a}_f\big)                   + \mathcal{L}\big(\widehat{\textbf{a}}_o,~ \textbf{a}_o\big)$  & 26.0 & 23.6  & 22.1 & 20.6 & 29.7 & 27.2 & 25.3 &  23.4\\

$ \mathcal{L}_{cyc}^{p} + \mathcal{L}_{cyc}^{s} + \mathcal{L}_{\textbf{S}} +  \mathcal{L}_{\textbf{P}}$               & \textbf{26.7}             & \textbf{24.3}             & \textbf{23.2}             & \textbf{21.7}                          & \textbf{30.8}             & \textbf{28.3}             & \textbf{26.1}             &  \textbf{25.8} \\
\toprule[1.2pt]

\end{tabular}
% }
\end{table*}
\begin{table*}[t]
% \vspace{1mm}
\captionsetup{width=.85\textwidth}
\caption{\textbf{Breakfast dataset: Comparison with the state-of-the-art methods.}The proposed method \textbf{A-ACT} outperforms the state-of-the-art approach LTCC~\cite{farha2020long} by a margin of $0.65\%$ for 20\% observation and approximately by $0.8\%$ for 30\% observation.}
\label{tab:procedural_results_bf}
\centering
% \resizebox{\columnwidth}{!}{%
\renewcommand{\arraystretch}{1.4}
\begin{tabular}{
% >{\centering\arraybackslash}p{0.5cm}
>{\raggedright\arraybackslash}p{2.4cm}|>{\centering\arraybackslash}p{0.6cm}>{\centering\arraybackslash}p{0.6cm}>{\centering\arraybackslash}p{0.6cm}>{\centering\arraybackslash}p{0.6cm}|>{\centering\arraybackslash}p{0.6cm}>{\centering\arraybackslash}p{0.6cm}>{\centering\arraybackslash}p{0.6cm}>{\centering\arraybackslash}p{0.6cm}}
% \begin{tabular}{l|cccc|cccc}
\toprule[1.2pt]
\multirow{2}{*}{\textbf{Methods}} & \multicolumn{4}{c|}{\textbf{20\%}}                                                                & \multicolumn{4}{c}{\textbf{30\%}}                                                            \\
\cline{2-9}
& 10\%                      & 20\%                     & 30\%                     & 50\%                       & 10\%                      & 20\%                      & 30\%                      & 
                               50\% \\
\toprule[1.2pt]
% \multicolumn{9}{l}{Breakfast} \\ \hline
% \parbox[t]{2.5mm}{\multirow{5}{*}{\rotatebox[origin=c]{90}{\textbf{Breakfast}}}} 

 TAB~\cite{sener2020temporal} & 24.2         & 21.1                      & 20.0                      & 18.1                                               & {\ul 30.4}                      & 26.3                      & 23.8                      & 21.2                      \\
UAAA~\cite{farha2019uncertainty}     & 16.7                 & 15.4                 & 14.5                & 14.2                 & 20.7                & 18.3                & 18.4                & 16.8  \\
Time-Cond.~\cite{ke2019time}                & 18.4                 & 17.2                & 16.4                & 15.8                & 22.8                & 20.4                & 19.6                & 19.8  \\

 LTCC~\cite{farha2020long}                                 & 25.9                     & 23.4                      & 22.4                      & 21.6                                               & 29.7                      & 27.4                     & 25.6                      &25.2 \\ \cline{1-9}
 \textbf{A-ACT} (ours)                  & \textbf{26.7}             & \textbf{24.3}             & \textbf{23.2}             & \textbf{21.7}                          & \textbf{30.8}             & \textbf{28.3}             & \textbf{26.1}             &  \textbf{25.8} \\
\toprule[1.2pt]
\end{tabular}
% }
% \vspace{-3mm}
\end{table*}

\noindent \textbf{Comparison with the state-of-the-art.} We compare our cycle-transformation model using all the appearance, optical flow and object features on the test sets provided by authors in~\cite{furnari2019would} to compare fairly against state-of-the-art-methods. To use all the three features we employ the late fusion strategy similar to what is used in~\cite{furnari2019would}.  Table~\ref{tab:test_results} assesses the performance evaluation of our proposed framework with state-of-the-art methods on the official test dataset. Our proposed approach outperforms the state-of-the-art RU-LSTM by consistent margin for all the tasks of action, noun and verb anticipation. Our approach shows improvement of 3\% and 2.36\% in terms of Top-5 accuracy for action anticipation for seen (S1) and unseen (S2) dataset, respectively. Also, the approach is very competitive with Action-Bank~\cite{wu2019long} and Ego-OMG~\cite{ dessalene2020egocentric} methods which uses superior features and additional features, respectively.\vspace{-2mm}

\subsection{Anticipation on Procedural Activities}
We evaluate the performance of anticipation models on procedural activities using different observation and prediction percentages. We conduct experiments to compare different anticipation models and compare our proposed framework with the state-of-the-art-methods.\smallskip
% We also conduct an ablation experiment to evaluate the contribution of different loss terms for anticipating procedural activities.
\begin{figure*}[!t]
     \centering
     \captionsetup[sub]{font=large}
     \begin{subfigure}[b]{0.46\textwidth}
         \centering
         \resizebox{0.96\textwidth}{!}{%
         \pgfkeys{
   /pgf/number format/.cd, 
    %   set decimal separator={.{\!}},
      set thousands separator={}
}
\pgfplotsset{
   every axis/.append style = {
      line width = 1pt,
      tick style = {line width=1pt}
   }
}

\begin{tikzpicture}

\begin{axis}[
    xlabel={Prediction (\%) with 20\% observation},
	ylabel={ Mean over Classes (MoC) accuracy (\%)},
	xmin=10, xmax=50,
	ymin=10, ymax=40,
	legend pos=north east,
    ymajorgrids=true,
    xmajorgrids=true,
    grid style=dashed,
    xtick={10, 20, 30, 40, 50},
]
% \begin{axis}[
% 	x tick label style={
% 		/pgf/number format/1000 sep=},
% 	ylabel=Year,
% 	enlargelimits=0.05,
% 	legend style={at={(0.5,-0.1)},
% 	anchor=north,legend columns=-1},
% 	ybar interval=0.1,
% ]
\addplot
	coordinates {(10, 33.1) (20, 26.9) (30, 21.2) (50, 14.6)};
\addplot 
	coordinates {(10, 34.0) (20, 27.4) (30, 21.5) (50, 14.9)};
\addplot+[brown, mark=triangle*, mark size=3.25pt]
    coordinates {(10, 35.4) (20, 29.6) (30, 22.5) (50, 16.1)};
\legend{SE, PV, A-ACT}
\end{axis}
\end{tikzpicture}
         }
         \caption{Anticipation with 20\% observation.}
         \label{fig:procedural_baselines_s50_a}
     \end{subfigure}
     \hfill
     \begin{subfigure}[b]{0.46\textwidth}
         \centering
         \resizebox{0.96\textwidth}{!}{%
         \pgfkeys{
   /pgf/number format/.cd, 
    %   set decimal separator={.{\!}},
      set thousands separator={}
}
\pgfplotsset{
   every axis/.append style = {
      line width = 1pt,
      tick style = {line width=1pt}
   }
}

\begin{tikzpicture}

\begin{axis}[
    xlabel={Prediction (\%) with 30\% observation},
	ylabel={ Mean over Classes (MoC) accuracy (\%)},
	xmin=10, xmax=50,
	ymin=10, ymax=40,
	legend pos=north east,
    ymajorgrids=true,
    xmajorgrids=true,
    grid style=dashed,
    xtick={10, 20, 30, 40, 50},
]
% \begin{axis}[
% 	x tick label style={
% 		/pgf/number format/1000 sep=},
% 	ylabel=Year,
% 	enlargelimits=0.05,
% 	legend style={at={(0.5,-0.1)},
% 	anchor=north,legend columns=-1},
% 	ybar interval=0.1,
% ]
\addplot
	coordinates {(10, 33.0) (20, 22.9) (30, 17.5) (50, 14.9)};
\addplot 
	coordinates {(10, 34.2) (20, 23.1) (30, 18.1) (50, 15.3)};
\addplot+[brown, mark=triangle*, mark size=3.25pt]
    coordinates {(10, 35.7) (20, 25.3) (30, 20.1) (50, 16.3)};
\legend{SE, PV, A-ACT}
\end{axis}
\end{tikzpicture}
         }
         \caption{Anticipation with 30\% observation.}
         \label{fig:procedural_baselines_s50_b}
     \end{subfigure}
     \caption{\textbf{Salad50 Dataset: Performance of different anticipation mechanisms} Similar to the Breakfast dataset, \textbf{PV} model outperforms \textbf{SE} for different predictions. With 20\% observation, the proposed \textbf{A-ACT} model outperforms the \textbf{SE} model by an average of $1.98\%$ and the \textbf{PV} model by an average margin of $1.45\%$. Our model also show improvement of $2.28\%$ over the semantic experience model and $1.68\%$ over the pattern visualization model for 30\% observation.}
     \label{fig:procedural_baselines_s50}
\end{figure*}
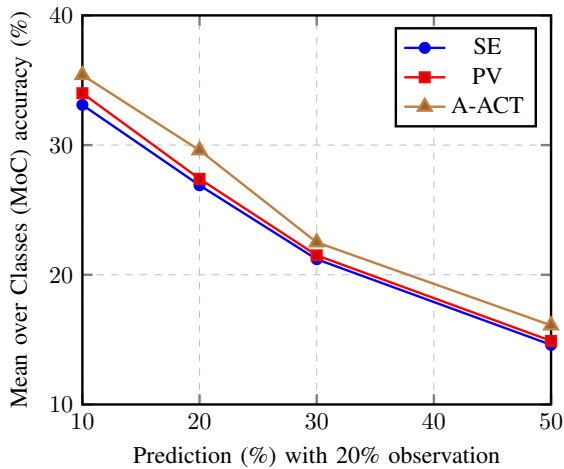
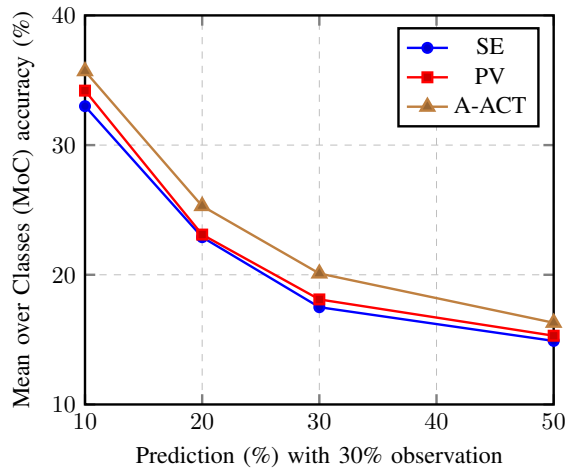
\begin{table*}[t]
\centering
% \vspace{1mm}
\captionsetup{width=.845\textwidth}
\caption{\textbf{50Salads: Impact of different loss terms.} Unlike the Breakfast dataset, in the experiment with 30\% observation, anticipation of the future activities at different prediction range using the proposed approach \textbf{A-ACT} does not improve as much as in the Breakfast dataset. Note: $\mathcal{L}_{\textbf{S}}$ and $\mathcal{L}_{\textbf{P}}$ consists of the action classification loss $\mathcal{L}\big(\widehat{\textbf{a}}_o,~ \textbf{a}_o\big)$ and action anticipation loss $\mathcal{L}\big(\widehat{	\textbf{a}}_f^{~p},~ \textbf{a}_f\big)$ as in row 2.}
\label{tab:ablation_loss_s50}
% \parbox{\columnwidth}{
% \resizebox{\columnwidth}{!}{%
\centering
\renewcommand{\arraystretch}{1.5}
\begin{tabular}{l|cccc|cccc}
% {
% % >{\centering\arraybackslash}p{0.5cm}
% >{\raggedright\arraybackslash}p{2.5cm}|>{\centering\arraybackslash}p{0.5cm}>{\centering\arraybackslash}p{0.5cm}>{\centering\arraybackslash}p{0.5cm}>{\centering\arraybackslash}p{0.5cm}|>{\centering\arraybackslash}p{0.5cm}>{\centering\arraybackslash}p{0.5cm}>{\centering\arraybackslash}p{0.5cm}>{\centering\arraybackslash}p{0.5cm}}
\toprule[1.2pt]
\multirow{2}{*}{\textbf{Model}} & \multicolumn{4}{c|}{\textbf{20\%}}                                                                & \multicolumn{4}{c}{\textbf{30\%}}                                                            \\
\cline{2-9}
& 10\%                      & 20\%                     & 30\%                     & 50\%                       & 10\%                      & 20\%                      & 30\%                      & 
                               50\% \\
\toprule[1.2pt]
%\parbox[t]{2.5mm}{\multirow{3}{*}{\rotatebox[origin=c]{90}{\textbf{50Salad}}}} 
% & 
$ \mathcal{L}_{cyc}^{p} + \mathcal{L}\big(\widehat{\textbf{a}}_f^{~p},~ \textbf{a}_f\big)$  & 33.9 & 25.7 & 19.6 & 13.1 & 32.8 & 22.3 & 16.5 &  13.7\\
% & 
$\mathcal{L}_{cyc}^{p} +  \mathcal{L}\big(\widehat{\textbf{a}}_f^{~p},~ \textbf{a}_f\big)                   + \mathcal{L}\big(\widehat{\textbf{a}}_o,~ \textbf{a}_o\big)$ & 34.8 & 27.1 & 21.0 & 14.4 & 33.2 & 22.6 & 17.2 & 14.8
\\
% &
$ \mathcal{L}_{cyc}^{p} + \mathcal{L}_{cyc}^{s} + \mathcal{L}_{\textbf{S}} +  \mathcal{L}_{\textbf{P}}$                  &  \textbf{35.4} &  \textbf{29.6} &  \textbf{22.5} &  \textbf{16.1} &  \textbf{35.7} &  \textbf{25.3}&  \textbf{20.1} &  \textbf{16.3}\\
\toprule[1.2pt]
\end{tabular}
% }
% }
\end{table*}
\begin{table*}[t]
% \vspace{4mm}
\captionsetup{width=.85\textwidth}
\caption{\textbf{Salad50 dataset: Comparison with the state-of-the-art methods.} We show about 1.1\% and 0.8\%  gain  over  the  state-of-the-art approach LTCC~\cite{farha2020long} in predicting 10\% in the future for 20\% and 30\% observation, respectively.}
% \vspace{0.4mm}
\label{tab:procedural_results_salad}
\centering
% \resizebox{\columnwidth}{!}{%
\renewcommand{\arraystretch}{1.3}
\begin{tabular}{
% >{\centering\arraybackslash}p{0.5cm}
>{\raggedright\arraybackslash}p{2.4cm}|>{\centering\arraybackslash}p{0.6cm}>{\centering\arraybackslash}p{0.6cm}>{\centering\arraybackslash}p{0.6cm}>{\centering\arraybackslash}p{0.6cm}|>{\centering\arraybackslash}p{0.6cm}>{\centering\arraybackslash}p{0.6cm}>{\centering\arraybackslash}p{0.6cm}>{\centering\arraybackslash}p{0.6cm}}
\toprule[1.2pt]
\multirow{2}{*}{\textbf{Methods}} & \multicolumn{4}{c|}{\textbf{20\%}}                                                                & \multicolumn{4}{c}{\textbf{30\%}}                                                            \\
\cline{2-9}
& 10\%                      & 20\%                     & 30\%                     & 50\%                       & 10\%                      & 20\%                      & 30\%                      & 
                               50\% \\
\toprule[1.2pt] 
 TAB~\cite{sener2020temporal}                                & - & - & - & - & - & - & - & - \\
 UAAA~\cite{farha2019uncertainty}  & 24.9         & 22.4                & 19.9                & 12.8                & 29.1                 & 20.5                 & 15.3                & 12.3 \\             
 Time-Cond.~\cite{ke2019time}   & 32.5         & 27.6                & 21.3                & 16.0                & 35.1                & 27.1                & 22.1                & 15.6 \\
 LTCC~\cite{farha2020long}                                   &34.8 &28.4 &21.8 &15.3 &34.4 & 23.7 & 19.0 & 15.9 \\
 \cline{1-9}
 \textbf{A-ACT} (ours)                  &  \textbf{35.4} &  \textbf{29.6} &  \textbf{22.5} &  \textbf{16.1} &  \textbf{35.7} &  \textbf{25.3}&  \textbf{20.1} &  \textbf{16.3}\\
\toprule[1.2pt]
\end{tabular}
% }
% \vspace{-2mm}
\end{table*}

\noindent \textbf{Breakfast dataset: Baselines.} We evaluate the performance of different anticipation models on Breakfast dataset by computing the anticipation accuracy for different observation and prediction percentage of the input video. Comparison of different anticipation models for 20\% and 30\% observation at different prediction horizons is shown in Figure~\ref{fig:procedural_baselines}. For all the experiments, our A-ACT model outperforms the semantic experience model and pattern visualization model the Breakfast dataset. Experiments also suggest that the pattern visualization model consistently outperforms the semantic experience model. The pattern visualization model reduces the error propagation in semantic anticipation from observed actions to future actions by first generating plausible future features and then performing anticipation on the features generated using translation model.\medskip
% \\[-1em]

\noindent \textbf{Breakfast dataset: Impact of different loss terms.} 
%\akash{Should we move these results to Supplemental?} 
Table~\ref{tab:ablation_loss_bf} presents the ablation for loss terms used for training our proposed framework on Breakfast dataset. We observe similar trend in performance for procedural activities datasets as in Epic-Kitchen dataset. It is interesting to observe that for 20\% observation the feature visualization loss contributes more for anticipating further in the time horizon. The improvement by adding recognition loss $\mathcal{L}\big(\widehat{\textbf{a}}_o,~ \textbf{a}_o\big)$ is only 0.7\% for 10\% and 20\% future prediction compared to 1.1\% improvement for 30\% and 50\% prediction. For 30\% observation, all the models show improvement of 4-5\% over 20\% observation performance. It can be attributed to more information available by increase in observation percentage from 20\% to 30\%.\smallskip

\noindent \textbf{Breakfast dataset: Comparison with the state-of-the-art.}
We compare the proposed cycle-transformation framework with the state-of-the-art methods on Breakfast and Salad50 dataset. Table ~\ref{tab:procedural_results_bf} compares the performance of our approach with state-of-the-art methods. All the approaches except ~\cite{farha2020long} follows a two-step approach by first identifying the activity in the observed frames and then use these labels to anticipate the future activities. Authors in~\cite{farha2020long} adopt the sequence-to-sequence approach for their framework. Unlike these methods our approach not only anticipates the action based on inferred activity labels in the observed frames but also synthesize probable future features and then perform anticipation. We also take advantage of the available future features and labels during training to enforce cycle-transformation in semantic label as well as feature space. 
It can be observed from the Table~\ref{tab:procedural_results_bf} that our approach outperforms all state-of-the-art approaches. As expected, performance gain of the near future 10\% prediction with 30\% observation is higher when compared to 20\% observation ($1.1\%$ vs $0.8\%$) over state-of-the-art LTCC~\cite{farha2020long} approach.\smallskip
% \amit{I do not understand the flow of this section. You mention 3 datasets, but then there is a section only on EPIC Kitchens, but then at some places, it mentions other datasets too. What dataset are the Procedural Activities on?}
%Comparison with state-of-the-art approaches is presented in Figure~\ref{fig:procedural_results_bf}. The anticipation performance with 20\% observation drops by 1.5\% when comparing the anticipation from prediction percentage of 30\% to 50\% . Interestingly, for the observation of 30\% the drop is merely 0.3\%.

% For the BreakFast dataset, refer , our proposed framework outperforms all the state-of-the-art methods. The anticipation performance with 20\% observation drops by 1.5\% when comparing the anticipation from prediction percentage of 30\% to 50\% . Interestingly, for the observation of 30\% the drop is merely 0.3\%. 

\noindent \textbf{Salad50 dataset: Baselines.} We additionally evaluate the performance of different anticipation models on 50Salads dataset. Figure~\ref{fig:procedural_baselines_s50} shows the comparison of different anticipation mechanism. When anticipating future with 20\% observation the proposed \textbf{A-ACT} model outperforms the semantic experience model by an average of $1.98\%$ and the pattern visualization model by an average margin of $1.45\%$. As expected the performance improvement of \textbf{A-ACT} with 30\% observation is $2.28\%$ over the semantic experience model and $1.68\%$ over the pattern visualization model. Also, as in experiments with different datasets, we observed the pattern visualization model consistently outperforms the semantic experience model in 50Salads dataset. The pattern visualization model reduces the error propagation in semantic anticipation from observed actions to future actions by first generating plausible future features and then performing anticipation on these features.\smallskip

\noindent \textbf{Salad50 dataset: Impact of different loss terms.} 
The ablation study on different loss terms used for training our proposed framework on 50Salads dataset is presented in Table~\ref{tab:ablation_loss_s50}. A similar trend in performance,  as in the EPIC-Kitchen and the Breakfast datasets, is observed on most of the experiments with the 50Salads dataset. The feature visualization loss contributes more for anticipating further in the time horizon. Interestingly, when using 20\% data as observation, the improvement by adding recognition loss $\mathcal{L}\big(\widehat{\textbf{a}}_o,~ \textbf{a}_o\big)$ is 0.9\% for 10\% and 1.4\% for 20\% future prediction compared to 1.4\% improvement for 30\% and 1.3\% for 50\% prediction. The average improvement for 20\% observation setting in 50Salads dataset is 1.3\% compared to Breakfast dataset where we see only 0.9\% average improvement. Unlike the Breakfast dataset, the experiment using 30\% data as observation, anticipation of future at different prediction range doesn't improve as much as the Breakfast dataset.\smallskip

\noindent \textbf{Salad50 dataset: Comparison with the state-of-the-art.} We perform additional experiments on Salad50 dataset to evaluate our approach. Comparison with the state-of-the art methods is presented in Table~\ref{tab:procedural_results_salad}. Our proposed approach \textbf{A-ACT} outperforms all state-of-the-arts approaches. As shown in Table~\ref{tab:procedural_results_salad}, our approach achieves about 1.1\% and 0.8\% gain over the state-of-the-art LTCC~\cite{farha2020long} approach for 20\% and 30\% observation time across different prediction lengths. As the approach in ~\cite{farha2020long} also employs the cycle-consistency but in label space, the performance gain using our approach can be attributed to the cycle-transformation in feature space along with the semantic label space. This shows that cycle-transformation in feature space is beneficial for anticipation. 
% For more results and ablation analysis on Salad50 dataset please refer the supplemental.

% ------------------ Conclusion
\section{Conclusion}

We study the task of action anticipation by leveraging the two anticipation models studied in human psychology for temporal anticipation. Specifically, we utilize the semantic experience and pattern visualization models to integrate the human anticipation mechanism in the proposed framework.  We present a framework \textbf{A-ACT} that combines both the semantic experience and pattern visualization model using cycle-transformations. Results suggest that cycle-transformation in semantic as well feature space helps learn the task of action anticipation better. It is observed that for many of our experiments the pattern visualization model slightly outperforms the semantic experience model. Experiments on standard datasets show efficacy of the proposed framework that utilizes combination of semantic experience and pattern visualization models using cycle-transformations against various state-of-the-art methods. 

\section*{Acknowledgment}
This work was partially supported by NSF grant 2029814.

\ifCLASSOPTIONcaptionsoff
  \newpage
\fi

\bibliographystyle{IEEEtran}
\bibliography{bib_files/VCG_Group_pubs, bib_files/VCG_Group_Refs, bib_files/local}

\begin{IEEEbiography}[{\includegraphics[width=1in,height=1.25in,clip,keepaspectratio]{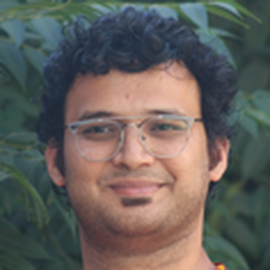}}]{Akash Gupta}
received his PhD and MS degree from the University of California, Riverside in Electrical and Computer Engineering in 2021. Previously, he received his Bachelor's degree in Electronics and Communications Engineering from Visvesvaraya National Institute of Technology, Nagpur in 2014. His research interests include computer vision, machine learning, video synthesis and enhancement, and video super-resolution.
\end{IEEEbiography}

\begin{IEEEbiography}[{\includegraphics[width=1in,height=1.25in,clip,keepaspectratio]{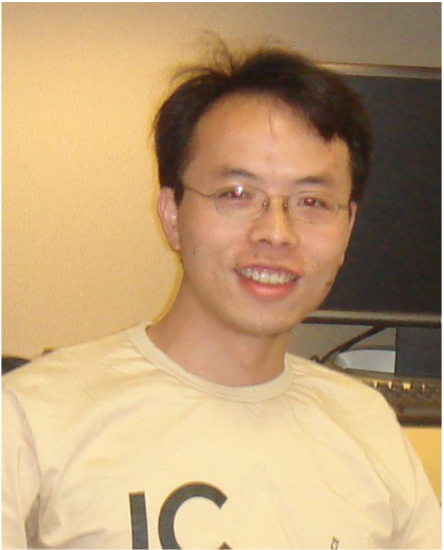}}]{Jingen Liu}
is a Principal Researcher at JD AI Research (JD.com Silicon Valley Research Center). Before joining JD.com, he was a Senior Computer Scientist at SRI International from 2011 to 2018, and a research fellow at University of Michigan, Ann Arbor from 2010 to 2011. He received his Ph.D. degree from the Center for Research in Computer Vision at University of Central Florida in 2009. His research is focused on computer vision, multimedia processing, medical image analysis and machine learning. His expertise lies in video analysis, human action recognition, multimedia event detection, semantic segmentation, and deep learning. He has published 50+ papers on prestigious AI conferences, such as CVPR, ICCV, ECCV, AAAI, and ACM Multimedia. As an outstanding young scientist, he was invited by NAE to present his work on video content analysis in the Japan-American Frontiers of Engineering Symposium 2012. He is the associate editors for international journals including IEEE Transaction on CSVT, Pattern Recognition, and IAPR Machine Vision \& Applications, and an area chair of IEEE \& detection evaluation workshops (2013 \& 2014).  
\end{IEEEbiography}

\vspace{-0.5cm}

\begin{IEEEbiography}[{\includegraphics[width=1in,height=1.25in,clip,keepaspectratio]{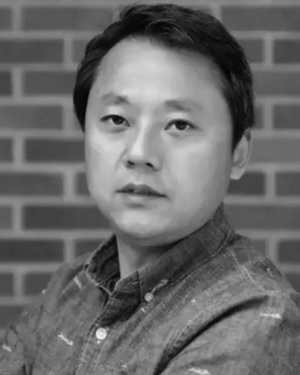}}]{Liefeng Bo} (Member, IEEE) received the Ph.D. degree from Xidian University, in 2007. He is currently a Distinguished Scientist and the Head of AI Lab at JD Finance. He was a Principal Scientist at Amazon, where he led a research team at Amazon Go for turning innovative research into products. Amazon Go is one of the most revolutionary innovations in the retail industry. It created a new grab-and-go shopping experience using computer vision, deep learning and sensor fusion technologies. He was a Postdoctoral Researcher successively at the Toyota Technological Institute at the University of Chicago and the University of Washington. His research interests are in machine learning, deep learning, computer vision, robotics, and natural language processing. He published more than 60 papers in top conferences and journals with more than 7000 Google Scholar citations. He won the National Excellent Doctoral Dissertation Award of China, in 2010, and the Best Vision Paper Award in ICRA 2011.
% (Based on document published on 26 January 2021).
\end{IEEEbiography}

\vspace{-0.5cm}

\begin{IEEEbiography}[{\includegraphics[width=1in,height=1.25in,clip,keepaspectratio]{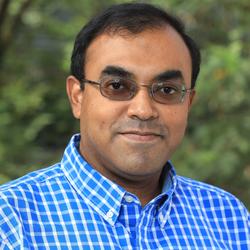}}]{Amit Roy-Chowdhury} received his PhD from the University of Maryland, College Park (UMCP) in Electrical and Computer Engineering in 2002 and joined the University of California, Riverside (UCR) in 2004 where he is a Professor and Bourns Family Faculty Fellow of Electrical and Computer Engineering, Director of the Center for Robotics and Intelligent Systems, and Cooperating Faculty in the department of Computer Science and Engineering. He leads the Video Computing Group at UCR, working on foundational principles of computer vision, image processing, and vision-based statistical learning, with applications in cyber-physical, autonomous and intelligent systems. He has published about 200 papers in peer-reviewed journals and conferences. He is the first author of the book Camera Networks: The Acquisition and Analysis of Videos Over~Wide Areas. His work on face recognition in art was featured widely in~the news media, including a PBS/National Geographic documentary and in The Economist. He is on the editorial boards of major journals and program committees of the main conferences in his area. His students have been first authors on multiple papers that received Best Paper Awards at major international conferences, including ICASSP and ICMR. He is a Fellow of~the IEEE and IAPR, received the Doctoral Dissertation Advising/Mentoring Award in 2019 from UCR, and the ECE Distinguished Alumni Award from UMCP.
\end{IEEEbiography}

\vspace{-0.5cm}

\begin{IEEEbiography}[{\includegraphics[width=1in,height=1.25in,clip,keepaspectratio]{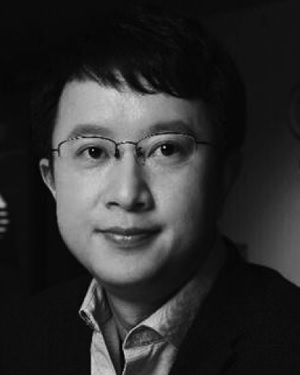}}]{Tao Mei} (Fellow, IEEE) received the B.E. and Ph.D. degrees from the University of Science and Technology of China, Hefei, China, in 2001 and 2006, respectively. He is currently the Vice President of JD.COM and the Deputy Managing Director of JD AI Research, where he also serves as the Director of the Computer Vision and Multimedia Laboratory. Prior to joining JD.COM in 2018, he was a Senior Research Manager with Microsoft Research Asia, Beijing, China. He has authored or coauthored over 200 publications (with 12 best paper awards) in journals and conferences, 10 book chapters, and edited 5 books. He holds over 25 U.S. and international patents. He is a Fellow of IAPR in 2016. He is the General Co-Chair of IEEE ICME 2019, and the Program Co-Chair of ACM Multimedia 2018, IEEE ICME 2015, and IEEE MMSP 2015. He is or has been an Editorial Board Member of IEEE Transactions on Image Processing, IEEE Transactions on Circuits and Systems for Video Technology, IEEE Transactions on Multimedia, ACM Transactions on Multimedia Computing, Communications, and Applications, and Pattern Recognition. He is a Distinguished Scientist of ACM in 2016 and a Distinguished Industry Speaker of the IEEE Signal Processing Society in 2017.
% (Based on document published on 30 July 2021).
\end{IEEEbiography}

\vfill

\end{document}